\pdfoutput=1
\documentclass[sigconf,nonacm]{acmart}
\usepackage{multirow}
\usepackage[table]{xcolor}
\usepackage{amsmath}
\usepackage{makecell}
\usepackage{adjustbox}
\AtBeginDocument{%
  }

\begin{document}

\title{MG-Thinker: Bi-Axial Self-Reflection for Multi-Image Reasoning Grounding}

\author{Heyu Huang}
\affiliation{%
  \institution{Huazhong University of Science and Technology}
  \city{Wuhan}
  \country{China}}

\author{Chi Chen}
\affiliation{%
  \institution{Tsinghua University}
  \city{Beijing}
  \country{China}}

\author{Zonghao Guo}
\affiliation{%
  \institution{Tsinghua University}
  \city{Beijing}
  \country{China}}

\author{Yuhua Li}
\affiliation{%
  \institution{Huazhong University of Science and Technology}
  \city{Wuhan}
  \country{China}}

\author{Maosong Sun}
\affiliation{%
  \institution{Tsinghua University}
  \city{Beijing}
  \country{China}}

\author{Ruixuan Li}
\affiliation{%
  \institution{Huazhong University of Science and Technology}
  \city{Wuhan}
  \country{China}}

\renewcommand{\shortauthors}{Huang et al.}

\begin{abstract}
Reinforcement learning (RL) has recently delivered substantial gains in multimodal reasoning, opening a promising route for fine-grained visual perception. Yet for \emph{multi-image reasoning grounding} (MRG), reasoning over real-world multi-image contexts toward pixel-precise localization, existing RL-based approaches overlook two characteristics intrinsic to this paradigm: a coarse-to-fine hierarchical reasoning pattern, and heterogeneously distributed task--sample difficulties. In this work, we present MG-Thinker, a post-training RL framework that advances a new MRG paradigm featuring such hierarchical reasoning, supported by a curated 25K MRG dataset with task-adaptive Chain-of-Thought (CoT) annotations that elicit multi-perspective evidence before conclusion. To remedy the heterogeneous task--sample difficulties, we further propose Bi-Axial DAPO (BiA-DAPO), which decomposes rollout advantages along an intra-group signal axis and an inter-group competence axis through two complementary mechanisms, both grounded on our defined candidate pool for stable group-level statistics. Extensive experiments show that MG-Thinker achieves state-of-the-art performance on multi-image reasoning grounding while consistently improving generalization across multi-image understanding and diverse multimodal benchmarks.
\end{abstract}


\maketitle

\section{Introduction}
\label{sec:intro}

\begin{figure}[!t]
  \centering
  \includegraphics[width=\columnwidth]{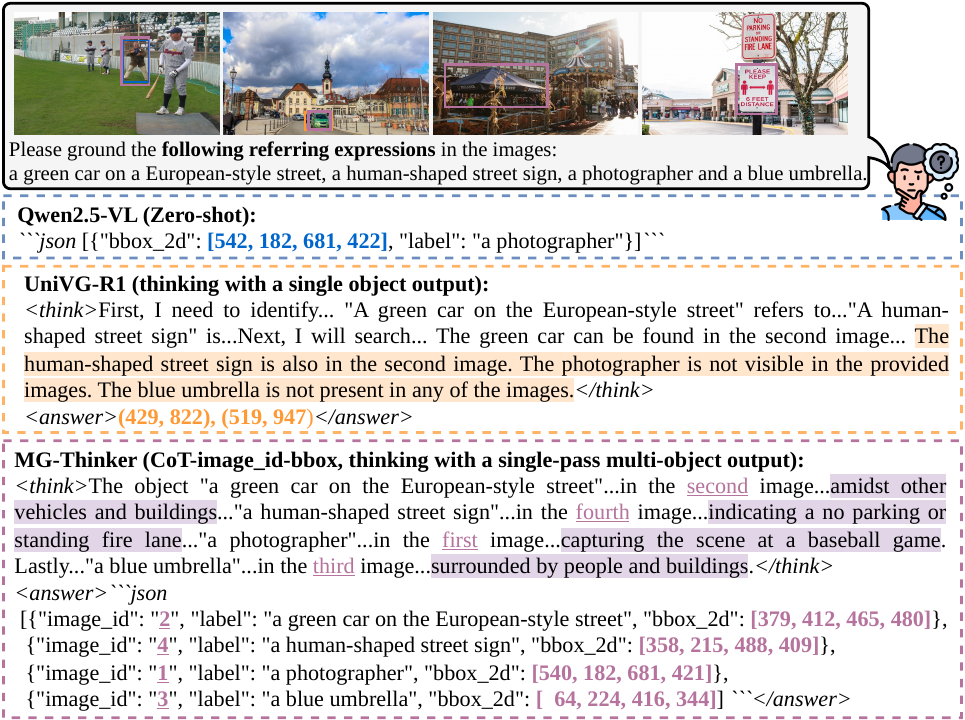}
  \caption{Illustrative case showing that multi-image reasoning grounding demands a \emph{coarse-to-fine hierarchical pattern} (CoT$\to$image-id$\to$bbox), absent in representative baselines, which MG-Thinker explicitly delivers through one-shot multi-image multi-object localization.}
  \label{fig1}
\end{figure}

Recent Multimodal Large Language Models (MLLMs)~\cite{lu2024deepseek,tong2024cambrian,chen2024internvl,bai2025qwen2,NEURIPS2023_6dcf277e} have delivered promising progress in multimodal perception, spanning modal recognition~\cite{liu2024improved,ye2023ureader,wang2024cogvlm,guo2024llava}, single-image grounding~\cite{zhang2024gpt4roi,peng2023kosmos,you2023ferret,chen2023shikra}, and multi-image understanding~\cite{li2024llava,jiang2024mantis,ye2024mplug}. However, real-world deployment demands more fine-grained perception that generalizes across sequential images, videos, or multi-view captures while producing pixel-level precise outputs. Recent works~\cite{li-etal-2025-migician,bai2025univg} have begun to focus multi-image grounding, expanding from single-image perception toward cross-image free-form grounding queries. However, existing MLLMs typically rely on end-to-end direct prediction and suffer from grounding drift, unstable output formats, and weak cross-image alignment. Meanwhile, o1-style slow thinking~\cite{jaech2024openai,guo2025deepseek,team2025kimi,xu2024llava} and R1-like multimodal reasoning~\cite{zhan2025vision,shen2025vlm,cao2025ground} have markedly advanced visual reasoning but mainly in single-image scenarios, directly transplanting such single-image CoT into multi-image localization yields unfaithful rationales that misroute attention and amplify localization errors. These limitations motivate \emph{Multi-image Reasoning Grounding} (MRG), a new paradigm that couples reasoning traces cross-image with precise pixel-level grounding.

\begin{figure*}[!t]
    \centering
    \begin{minipage}[t]{0.52\textwidth}
        \centering
        \includegraphics[width=0.85\linewidth]{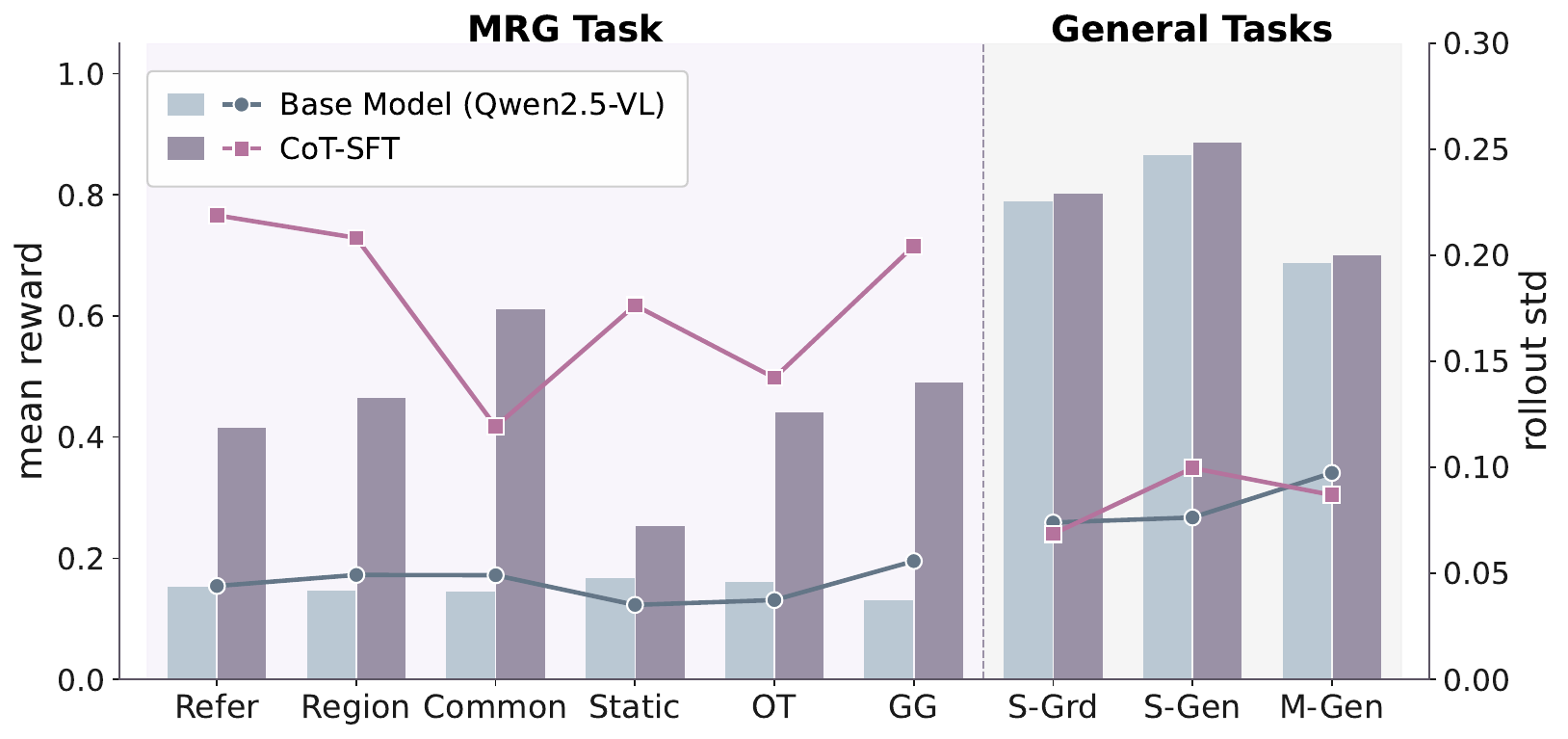}
    \end{minipage}
    \hfill
    \begin{minipage}[t]{0.43\textwidth}
        \centering
        \includegraphics[width=0.85\linewidth]{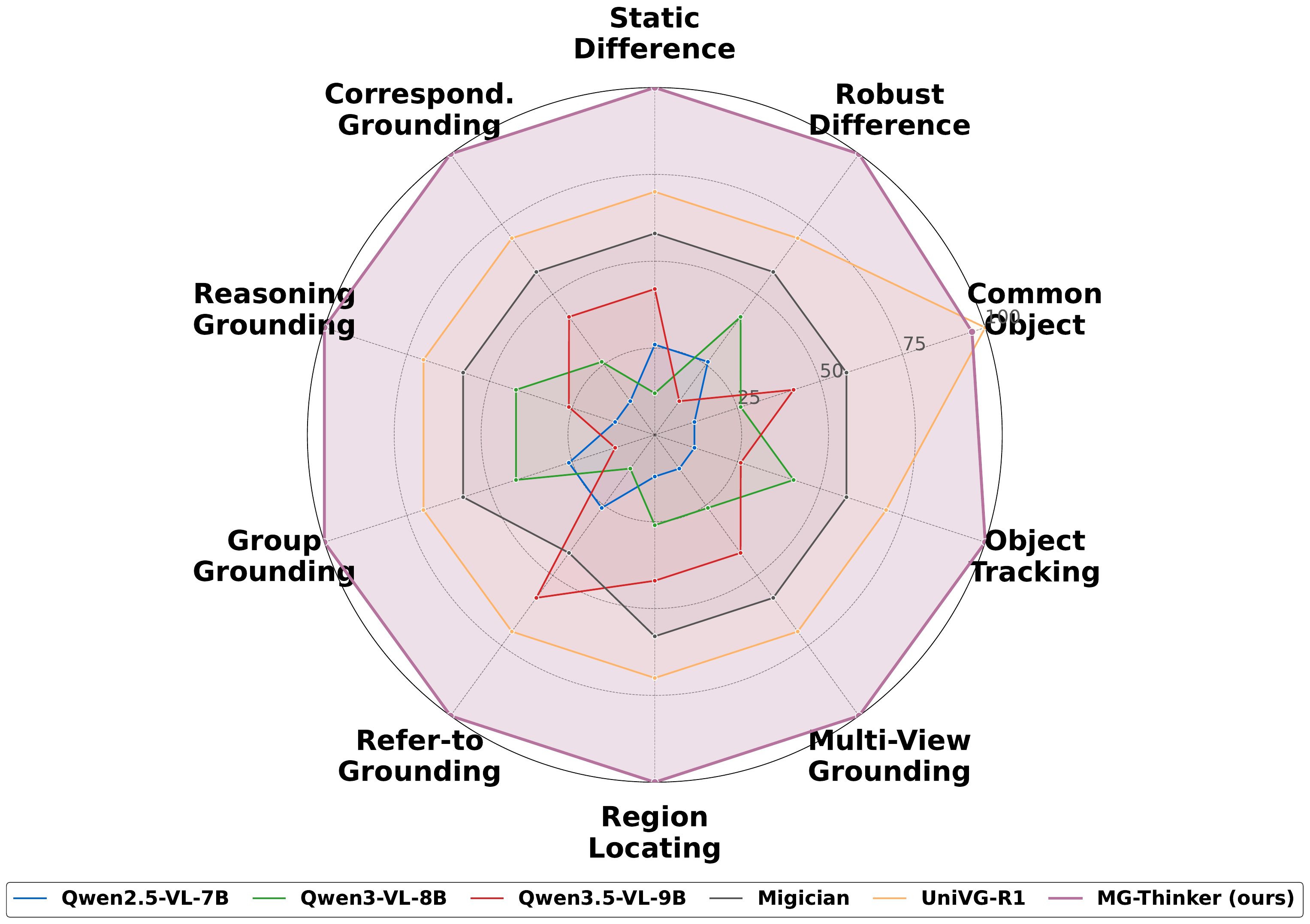}
    \end{minipage}
    \caption{(\textbf{Left}) Task--sample difficulty heterogeneity profile of MRG vs.\ general tasks: per-task mean reward (bars) and within-sample rollout-std (lines) over the base model (Qwen2.5-VL) and the CoT-SFT model; S-Grd, S-Gen, M-Gen denote single-image grounding, single-image and multi-image understanding tasks, respectively. (\textbf{Right}) Per-task accuracy radar over the MIG-Bench.}
    \label{fig2}
\end{figure*}
This paradigm presents two MRG-specific challenges that prior work largely overlooks. First, MRG inherently demands a coarse-to-fine hierarchical pattern, progressively unfolding from semantic-level reasoning to image-level routing to region-level localization, as shown in Fig.~\ref{fig1}, unlike generic tasks where templated reasoning or direct prediction suffices. Second, during post-training, MRG further exhibits pronounced task-sample difficulty heterogeneity. As shown in Fig.~\ref{fig2} (left), the base model (Qwen2.5-VL) trails far below the three general tasks on MRG in both mean reward and rollout-std, confirming that MRG as a whole is substantially harder than these general tasks. In contrast, the CoT-SFT model (the fine-tuned model before RL, detailed in Sec.~\ref{sec:data_curation}) swings widely across MRG sub-tasks while remaining stable on general tasks, jointly evidencing inter-task and sample-level difficulty heterogeneity.

Building on these challenges, we present MG-Thinker, an RL-based framework that addresses both characteristics through complementary data and algorithm designs. \textit{Data side.} To instill the coarse-to-fine hierarchy, we restructure MGrounding-630K and curate a 25K MRG-specific CoT dataset with task-adaptive cue prompts, so that multi-image evidence guides reasoning and reasoning, in turn, sharpens localization. This CoT design further folds in one-shot multi-image multi-target grounding from the start, addressing a joint-localization scenario that both direct-grounding and reasoning-based baselines noticeably struggle with. \textit{Algorithm side.} Heterogeneous task-sample difficulties induce two recurring training instabilities, \emph{drifting advantage-signal sparsity} from rollout groups with near-zero reward variance and \emph{competence-difficulty misalignment} from groups with widely different mean rewards. We thus propose Bi-Axial DAPO (BiA-DAPO), which stabilizes RL post-training through two simple group-level mechanisms: \emph{Group-Informativeness Assessment} (GIA) selects informative rollout groups via within-group reward variance to relieve the drifting sparsity, while \emph{Cascaded-Reward Stratification} (CRS) schedules updates over retained groups by group-level mean reward to match the policy's evolving competence, both grounded on a shared candidate pool for stable group-level statistics. This bi-axial view treats the group mean and within-group variance as statistically orthogonal signals that carry independent information about competence and informativeness, so a rollout group meaningfully contributes to the policy update only when well-positioned on both axes. As shown in Fig.~\ref{fig2} (right), MG-Thinker sets a new state of the art on MIG-Bench~\cite{li-etal-2025-migician} while extending consistent gains to broader multi-image and multimodal benchmarks.
\begin{figure*}[t]
  \centering
  \includegraphics[width=1.0\linewidth]{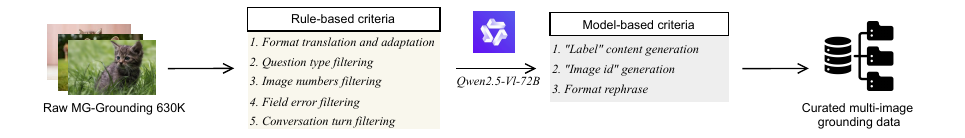}
  \caption{Illustration of the multi-image grounding data pre-processing pipeline, including rule-based criteria and processing operations derived using the multimodal large language model Qwen2.5-VL-72B.}
  \label{fig4}
\end{figure*}

Overall, our main contributions can be summarized as follows:
\begin{itemize}
\item We present MG-Thinker, a post-training RL framework that targets two MRG-specific characteristics overlooked by prior work, a coarse-to-fine hierarchical reasoning--grounding pattern, and heterogeneously distributed task--sample difficulties.
\item Data-side, we restructure MGrounding-630K and distill a 25K MRG-specific CoT dataset with task-adaptive cue prompts that instills the coarse-to-fine hierarchy. Algorithm-side, we propose Bi-Axial DAPO (BiA-DAPO), which improves group-relative RL by combining \emph{Group-Informativeness Assessment} (reward-variance-based group selection) with \emph{Cascaded-Reward Stratification} (reward-mean-based competence scheduling), both grounded on a shared candidate pool for stable group-level statistics.
\item Extensive experiments show that MG-Thinker achieves state-of-the-art performance on multi-image reasoning grounding while consistently generalizing across multi-image understanding, single-image grounding, and diverse multimodal benchmarks.
\end{itemize}
\section{Related Work}
\noindent\textbf{Visual Grounding.} Visual grounding\cite{zhang2024gpt4roi,peng2023kosmos,zhan2024griffon,wu2025f,you2023ferret,chen2023shikra,rasheed2024glamm} has evolved from referring-expression localization in single images, as in the RefCOCO series\cite{yu2016modeling,mao2016generation}, to reasoning-intensive instruction grounding (\emph{e.g.}, LISA-Grounding~\cite{lai2024lisa}), and further extends to free-form multi-image grounding paradigms closer to real-world applications. However, existing MLLMs remain limited in multi-image fine-grained grounding due to insufficient cross-image association precision~\cite{zhang2025cmmcot,fu2025hidden} and the lack of cross-image reasoning capability for free-form queries.

\noindent\textbf{Reinforcement Learning-based Reasoning.} MLLM reasoning has shown strong gains on visual reasoning~\cite{lu2023mathvista,wang2024measuring}. Prior works mostly construct CoT data with explicit reasoning steps~\cite{hu2024visual,shao2024visual,fan2025grit,zhan2025understand,li2024vocot} for SFT. With R1-like models, RL has emerged as a post-training paradigm for eliciting reasoning~\cite{liu2025visionreasoner}, mostly built on GRPO with rule-based rewards. Recent improvements span data construction, staged training, and reward/loss design~\cite{xiao2025fast}. VL-Rethinker~\cite{wang2025vl} replays high-value samples to address advantage vanishing, while AdaRFT~\cite{shi2025efficient} adaptively offline schedules sample difficulty against the policy's running rewards.

\section{Proposed Method}
\subsection{Overview}
Multi-image reasoning grounding (MRG) localizes target regions across an image set while producing an explicit cross-image reasoning trace. Given an image set $V{=}\{I_1,\ldots,I_m\}$, a prompt $P$ specifying intent and output format, and an optional reference $R$, the model first generates a reasoning trace $S$ and then outputs $n$ grounded predictions $\{(b_k,i_k)\}_{k=1}^{n}$, where each bounding box $b_k$ is indexed by an image identifier $i_k$. Formally,
\begin{equation}
\!\!O = \big\{S,\,\{(b_k,i_k)\}_{k=1}^{n}\big\} =
\begin{cases}
\mathcal{M}_{\text{sg}}(V,P), \\
\mathcal{M}_{\text{rg}}(V,P,R).
\end{cases}
\end{equation}
\textit{Spontaneous grounding (sg)} performs localization without an explicit reference, relying solely on $P$ to infer and search the target across the image set. \textit{Referential grounding (rg)} additionally consumes $R$ (a textual mention or visual exemplar), demanding robust cross-image correspondence and precise localization. All sub-tasks formulation, composition and properties detailed in Appendix~\ref{app:mig-tasks}.

\subsection{Data Curation and MRG-Specific CoT Activation}
\label{sec:data_curation}
For Supervised Fine-Tuning (SFT), we contribute a high-quality data curation pipeline that activates fine-grained perception capabilities in multi-image scenarios. As illustrated in Fig.~\ref{fig4}, we systematically reconstruct MGrounding-630K~\cite{li-etal-2025-migician}, a multi-image grounding corpus spanning six in-domain sub-tasks, into a refined training set adapted to Qwen2.5-VL~\cite{bai2025qwen2}, through a pipeline that combines rule-based filtering with model-assisted rewriting. Rule-based filtering operates over question types, image quantity and resolution, and dialogue turns to remove homogeneous, erroneous, over-resolution, and redundant-dialogue samples, while reformulating special tokens and JSON structures to match the target output format. For content that cannot be automatically rewritten, we employ Qwen2.5-VL-72B~\cite{wang2024qwen2} to regenerate label content from joint image-text context, introduce image-id key-value pairs for target image indices, and resample to maintain category balance across sub-tasks. The pipeline distills MGrounding-630K into 480K high-quality samples, from which 320K are drawn for stage-one SFT to establish a solid grounding-activation foundation before reasoning enhancement.

The next stage aims to activate MRG-specific reasoning by distilling task-adaptive CoT demonstrations that elicit cross-image evidence-seeking before localization, so that multi-image evidence guides reasoning and reasoning, in turn, sharpens grounding precision. We categorize the remaining 160K instances into four MRG-specific task types, visual comparative, spatial perception, temporal perception, and visual semantic/logical association, each of which corresponds to a distinct cross-image cue pattern that MRG hinges on and thus admits its own task-adaptive prompts (templates in Appendix~\ref{app:templates}), and prompt Qwen2.5-VL-72B to progressively analyze cross-image visual and textual evidence. Task-oriented operators (comparison, searching, observation, tracking, association) are injected to encourage step-wise evidence-seeking, while multi-image multi-object reasoning grounding is folded in from the beginning, enabling the model to reason once and then localize multiple targets across images in a single inference pass. The thinking process is constrained against revealing final answers or ground-truth bboxes, ensuring genuinely explanatory rationales rather than answer-leaking traces that RL would later exploit as shortcuts.

Two-step post-processing then ensures rationale rationality and utility: rule-based filtering removes pseudo-reasoning samples whose conclusions appear up-front, and IoU-improvement validation retains samples meeting one of three criteria, (i) correct direct prediction with $\geq$10\% IoU gain after CoT, (ii) incorrect direct prediction corrected by CoT, or (iii) incorrect direct prediction with $\geq$20\% IoU gain after CoT. Together, the pseudo-reasoning filter guards against answer-leaking rationales while the IoU-improvement criteria retain only CoT traces that demonstrably improve grounding, so downstream RL always inherits a cold-start prior that is both reasoning-faithful and grounding-effective. The pipeline yields 25K MRG cold-start samples, each formatted with a \textless\textit{think}\textgreater{}\textless\textit{/think}\textgreater{} reasoning trace followed by a \textless\textit{answer}\textgreater{}\textless\textit{/answer}\textgreater{} JSON block, forming the CoT-SFT initialization from which BiA-DAPO's RL post-training begins.
\begin{figure*}[t]
  \centering
  \includegraphics[width=0.95\linewidth]{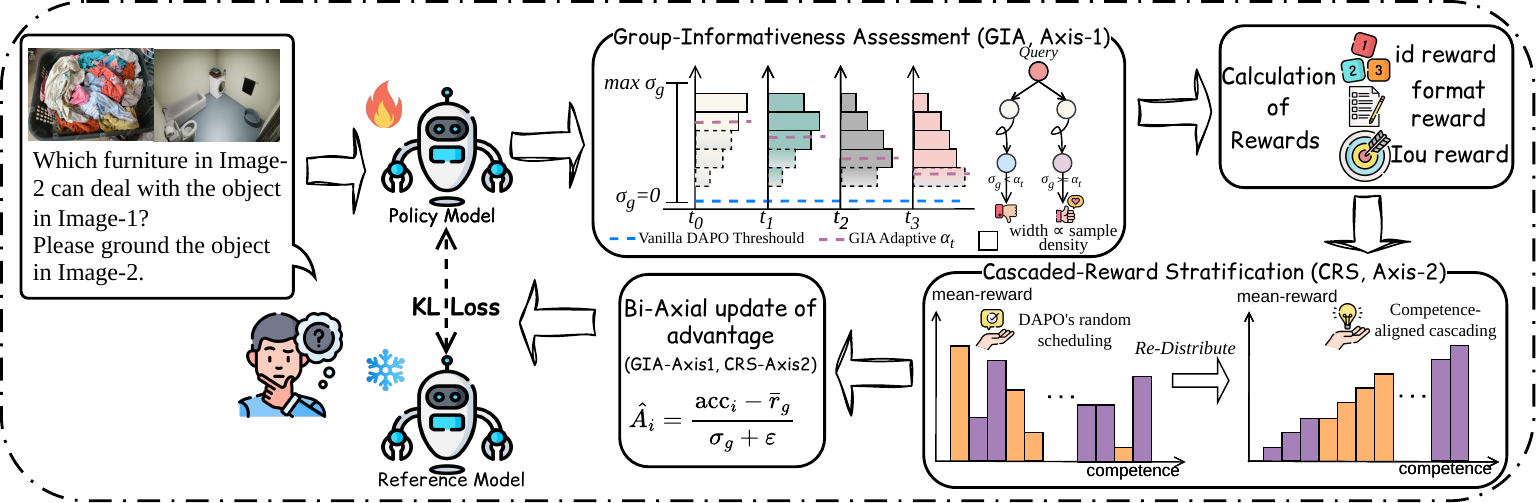}
  \caption{Overview of the proposed Bi-Axial DAPO (BiA-DAPO) for MRG. Upper: GIA scores each group along the intra-group signal axis to relieve drifting advantage-signal sparsity. Lower: CRS schedules updates over the GIA-retained groups along the inter-group competence axis to remedy competence-difficulty misalignment. Both operate on a shared candidate pool for stable group-level calculations.}
  \label{fig:biadapo}
\end{figure*}
\subsection{BiA-DAPO for Reasoning Enhancement}
\label{sec:biadapo}
\looseness=-1
In preliminary RL experiments on MRG, we observe two recurring training instabilities. First, many rollout groups exhibit near-zero reward variance and thus contribute vanishing group-relative advantage signals, effectively wasting the compute spent generating them. This sparsity drifts as the policy improves, since MRG's coarse-to-fine ladder ``CoT$\to$image-id$\to$bbox'' undergoes layer-wise saturation under AND-aggregated rewards, with each successive layer entering saturation as the previous one is mastered. Second, rollout groups with widely different mean rewards are mixed in the same update, causing the policy to oscillate between already-mastered and overly difficult samples. These two issues correspond, respectively, to \emph{drifting advantage-signal sparsity} and \emph{competence-difficulty misalignment}, two failure modes that a single group-level statistic cannot disentangle from one another.

Motivated by these observations, we propose BiA-DAPO, which uses two simple group-level statistics to stabilize RL training: \emph{Group-Informativeness Assessment} (GIA) leverages within-group reward variance to retain rollout groups whose gradients are informative and non-vanishing, while \emph{Cascaded-Reward Stratification} (CRS) uses group-level mean reward to schedule the retained groups from competence-mature to competence-developing bands, so each gradient step is delivered where the current policy can benefit from it most. The two mechanisms act on the same rollout groups but on distinct, complementary group-level statistics, jointly forming an advantage-grounded update stream that neither mechanism can produce on its own.

\noindent
\textbf{Bi-axial decomposition of advantage.} The DAPO group-relative advantage of rollout $i$ takes the form $\hat{A}_i = (\mathrm{acc}_i - \bar{r}_g)/(\sigma_g+\varepsilon)$, where $\bar{r}_g$ and $\sigma_g$ are the within-group mean and standard deviation of accuracy. The centring term encodes a group's \emph{competence position} along the inter-group competence axis (Axis-2), while the denominator encodes its \emph{informativeness} along the intra-group signal axis (Axis-1). The two terms are statistically orthogonal and carry independent information, a group thus yields a non-trivial policy gradient only when well-positioned on \emph{both} axes simultaneously. The two pathologies thus reside one per axis (sparsity on the signal axis, misalignment on the competence axis), and BiA-DAPO addresses each through a dedicated mechanism, all grounded on a shared candidate pool for stable group-level statistics.

\noindent
\textbf{Group-Informativeness Assessment (GIA).} GIA scores each group along the signal axis by its within-group reward variance $\sigma_g^2$, which reflects the magnitude of the advantage signal: a group whose rollouts receive near-identical rewards yields $\sigma_g{\approx}0$ and contributes negligible gradient information. We apply a \emph{variance-driven sampling gate} whose informativeness threshold linearly decays as training progresses: early in training the gate admits only the highest-variance groups to concentrate updates on the most informative samples, while later in training it progressively accepts lower-variance groups as the policy stabilizes. As shown in Fig.~\ref{fig:biadapo} (upper), color-coded nodes denote groups assessed against the current threshold. This adaptive gating replaces DAPO's all-or-nothing rejection and directly relieves advantage-signal sparsity by selectively excluding uninformative samples. The empirical training dynamics of this variance-driven gate, together with its effect on the group-level $\sigma_g$ distribution and the $R_{\text{IoU}}$ outcome composition across training phases, are analyzed in Appendix~\ref{app:dynamics}.

\noindent
\textbf{Cascaded-Reward Stratification (CRS).} CRS operates along the competence axis by stratifying the GIA-retained groups into ordered competence bands on the between-group mean reward $\bar{r}_g$, from high (competence-mature) to low (competence-developing). The policy is then updated band by band along this ordering, as shown in Fig.~\ref{fig:biadapo} (lower), forming a \emph{reward-driven training schedule} that naturally transitions from easy to hard samples so that each gradient step consumes groups whose difficulty matches the policy's evolving capability. The target competence band thus emerges \emph{dynamically} from reward statistics without external difficulty labels, directly remedying competence-difficulty misalignment over the grouped rollouts. Empirical evidence for this cascade, in particular the diagonal migration of retained groups on the joint $(\bar{r}_g, \sigma_g)$ plane from low-competence high-spread toward high-competence moderate-spread regions, is further presented in Appendix~\ref{app:dynamics}.

\noindent
\textbf{Stable group-level statistics.} Reliable GIA and CRS decisions require stable axis calculations, since both $\sigma_g$ and $\bar{r}_g$ become noisy when computed from too few groups. Therefore, we repurpose DAPO's standard ``generation batch'' as our candidate pool for each update step, whose existing size already suffices for group-level statistics to remain stable under bootstrap resampling, without any further enlargement on our part. The integration of GIA and CRS over this candidate pool constitutes BiA-DAPO, which jointly relieves drifting advantage-signal sparsity and remedies competence-difficulty misalignment in a principled, advantage-grounded manner, without introducing any auxiliary loss, external difficulty labels, or generation-batch enlargement beyond the two group-level statistics themselves.

\subsection{Reward Modeling and Training Objective}
\label{sec:reward}
\looseness=-1
Following R1-like approaches, we employ a deterministic, coarse-to-fine reward tailored to MRG: a format reward enforces the reasoning-to-answer protocol, while the accuracy reward decomposes into image-id and IoU components that first validate source image(s) and then assess box localization. We refer to $R_{\text{acc}}{=}R_{\text{id}}{+}R_{\text{IoU}}$ as the \emph{post-format accuracy} (PFA), which empirically dominates the RL-active signal after CoT-SFT, also motivating our bi-axial treatment.

\noindent
\textbf{Format Reward.} Outputs are checked for \textless\textit{think}\textgreater{}/\textless\textit{answer}\textgreater{} tags and the JSON block, with $0.25$ per correctly-paired tag (else $0$).

\noindent
\textbf{Image-id Reward.} As the coarse-grained accuracy component, we first verify image-level correctness before proceeding to localization evaluation. For multi-image scenarios, Hungarian matching aligns predicted with ground-truth JSON blocks by image-id, while single-object cases use direct matching. The reward is $R_{\text{id}}{=}1$ if and only if image-id matches, otherwise $0$, in which case the subsequent IoU evaluation is skipped.

\noindent
\textbf{IoU Reward.} As the fine-grained component, we evaluate target-level bounding-box precision only after a successful image-id match, using a dual-threshold mechanism (high $\tau_h$, low $\tau_l$) that encourages high-precision localization while preventing reward hacking through degenerate bounding boxes. For multi-image multi-object inputs, each target undergoes one $R_{\text{id}}$ and one $R_{\text{IoU}}$ computation, with per-target rewards summed sequentially:
\begin{equation}
R_{\text{total}} \!=\! R_{\text{format}} + R_{\text{acc}},\;\; R_{\text{acc}} \!=\! R_{\text{id}} + R_{\text{IoU}}
\end{equation}
\begin{equation}
R_{\text{IoU}} \!=\! \begin{cases}
1, & \text{IoU}(B,\tilde{B}) \geq \tau_h \\
\text{IoU}(B,\tilde{B}), & \tau_l \!\leq\! \text{IoU} \!\leq\! \tau_h \\
0, & \text{IoU}(B,\tilde{B}) \leq \tau_l
\end{cases}
\end{equation}
$B$ and $\tilde{B}$ are predicted and ground-truth boxes; $\tau_h{=}0.9$, $\tau_l{=}0.1$, with linear interpolation between thresholds.

\noindent
\textbf{Training Objective.} For the DAPO training objective, let question $q$ paired with answer $a$, and ${o_1, \ldots, o_G}$ denote a group of sampled outputs from the old policy model $\pi_{\theta_{\text{old}}}$. The optimization objective of policy model $\pi_{\theta}$ is formally defined as:
\begin{multline}
\setcounter{equation}{5}
\mathcal{J}_{\text{DAPO}}(\theta) = \mathbb{E}_{(q,a)\sim \mathcal{D}, \{o_i\}_{i=1}^G\sim \pi_{\theta_\text{old}}(\cdot\mid q)} \Bigg[\frac{1}{\sum_{i=1}^{G}|o_i|}\sum_{i=1}^{G}\sum_{t=1}^{|o_i|} \\
\min \Big( r_{i,t}(\theta) \hat{A}_{i,t}, \text{clip} \Big(r_{i,t}(\theta), 1 - \varepsilon_{low}, 1 + \varepsilon_{high} \Big) \hat{A}_{i,t} \Big) \Bigg] \tag{5}
\label{eq:dapoloss}
\end{multline}
where $\varepsilon_{\text{low}}$ and $\varepsilon_{\text{high}}$ denote the lower and upper clipping parameters, respectively. The clipping function bounds reward ratios to $[1-\varepsilon_{\text{low}}, 1+\varepsilon_{\text{high}}]$, preventing excessive policy updates. The constraint $0<|\{o_i \mid \texttt{is\_equivalent}(a, o_i)\}|<G$ ensures that among $G$ generated outputs, at least one but not all sequences are equivalent to target action $a$, maintaining diversity while enabling policy learning from both positive and negative examples. The DAPO training formulation details are in Appendix~\ref{app:dapo}.

The term $r_{i,t}$ represents the probability ratio at timestep $t$ for the $i$-th output sequence under current policy $\pi_\theta$, while $\hat{A}_{i,t}$ denotes the advantage computed based on group rewards $\{R_1, \ldots, R_G\}$:
\begin{equation}
\begin{aligned}
    r_{i,t}(\theta)=\frac{\pi_{\theta}(o_{i,t} \mid q, o_{i,<t})}{\pi_{\theta_{\text{old}}}(o_{i,t} \mid q,o_{i,<t})},\quad\hat{A}_{i,t} = \frac{R_i - \text{mean}(\{R_i\}_{i=1}^G)}{\text{std}(\{R_i\}_{i=1}^G)} 
\end{aligned}
\label{eq:advantage_calculation}
\end{equation}

\begin{table*}[t]
\centering
\caption{Performance comparison on MIG-Bench. OT, MV, GG and Co-Re respectively mean object tracking, multi-view, grouped and correspondence grounding. $^{\dagger}$ denotes our fair re-implementation of existing RL baselines.}
\begin{adjustbox}{max width=\linewidth}
\setlength{\tabcolsep}{4pt}
\renewcommand{\arraystretch}{0.85}
\small
\begin{tabular}{l|cc|c|cccc|c|cc|c}
\toprule
\multirow{4}{*}{\textbf{Models}} & \multicolumn{3}{c|}{\textbf{Spontaneous Grounding}} & \multicolumn{7}{c|}{\textbf{Referential Grounding}} & \multirow{4}{*}{\textbf{AVG}} \\ 
\cmidrule(lr){2-4} \cmidrule(lr){5-11} 
\multicolumn{1}{c}{} & \multicolumn{2}{|c|}{\textbf{Difference}} & \textbf{Similarity} & \multicolumn{4}{c|}{\textbf{Visual Reference}} & \textbf{Textual} & \multicolumn{2}{c|}{\textbf{Visual+Textual}} & \\ 
\cmidrule(lr){2-3} \cmidrule(lr){4-4} \cmidrule(lr){5-8} \cmidrule(lr){9-9} \cmidrule(lr){10-11}
\multicolumn{1}{c|}{} & \textbf{Static} & \textbf{Robust} & \multicolumn{1}{c|}{\textbf{Common}} & \textbf{OT} & \textbf{MV} & \textbf{Region} & \textbf{Refer} & \multicolumn{1}{c|}{\textbf{GG}} & \textbf{Reason} & \multicolumn{1}{c|}{\textbf{Co-Re}} & \\
\midrule
\rowcolor{gray!20}
\multicolumn{12}{c}{\textbf{70B-Scale MLLMs}}\\
\midrule
LLaVA-OV-72B~\cite{li2024llava} & 13.26 & 5.34 & 26.84 & 12.91 & 7.64 & 2.14 & 17.83 & 21.60 & 11.88 & 8.55 & 13.65 \\
InternVL2-76B~\cite{chen2024internvl} & 15.91 & 10.64 & 36.40 & 30.73 & 20.83 & 5.74 & 46.46 & 41.28 & 32.67 & 26.50 & 26.72 \\
InternVL3-78B~\cite{zhu2025internvl3} & 10.04 & 9.57 & 24.12 & 27.08 & 14.58 & 10.44 & 50.51 & 38.08 & 45.54 & 17.09 & 24.71 \\
Qwen2-VL-72B~\cite{wang2024qwen2} & 46.12 & 46.81 & 64.46 & 26.73 & 22.57 & 18.62 & 33.33 & 62.53 & 50.50 & 17.09 & 38.88 \\
Qwen2.5-VL-72B~\cite{bai2025qwen2} & 43.75 & 46.81 & 69.98 & 34.32 & 29.17 & 8.31 & 62.63 & 59.92 & 66.34 & 41.03 & 46.23 \\
\midrule 
\rowcolor{gray!20}
\multicolumn{12}{c}{\textbf{7B-Scale MLLMs}}\\
\midrule
Mantis~\cite{jiang2024mantis} & 1.52 & 0.00 & 3.31 & 12.18 & 2.08 & 1.00 & 1.01 & 10.02 & 0.00 & 0.85 & 3.20 \\
LLaVA-OV-7B~\cite{li2024llava} & 6.06 & 3.19 & 3.43 & 0.18 & 1.04 & 1.08 & 9.09 & 15.43 & 6.93 & 0.85 & 4.73 \\
MiniCPM-V-2.6~\cite{yao2024minicpm} & 14.58 & 2.13 & 14.34 & 9.82 & 6.25 & 1.75 & 11.11 & 10.02 & 2.97 & 2.56 & 7.55 \\
mPLUG-Owl3~\cite{ye2024mplug} & 18.56 & 6.38 & 34.93 & 8.55 & 7.64 & 2.41 & 7.07 & 22.85 & 9.09 & 5.98 & 12.35 \\
InternVL2-8B~\cite{chen2024internvl} & 6.92 & 7.45 & 25.49 & 20.73 & 9.72 & 3.49 & 28.28 & 30.26 & 17.82 & 9.40 & 15.96 \\
InternVL3-8B~\cite{zhu2025internvl3} & 23.67 & 14.89 & 47.99 & 14.84 & 6.94 & 12.13 & 7.07 & 34.87 & 16.83 & 2.56 & 18.18 \\
Qwen2-VL-7B~\cite{wang2024qwen2} & 27.84 & 38.30 & 19.36 & 20.73 & 11.81 & 25.95 & 23.23 & 58.52 & 48.51 & 11.97 & 28.62 \\
Qwen2.5-VL-7B~\cite{bai2025qwen2} & 29.92 & 22.45 & 16.36 & 6.94 & 3.66 & 15.05 & 35.67 & 63.51 & 4.03 & 3.42 & 20.10 \\
Qwen3-VL-8B~\cite{bai2025qwen3} & 29.36 & 24.47 & 58.83 & 28.00 & 17.36 & 23.44 & 33.33 & 70.93 & 32.99 & 10.25 & 32.90 \\
Qwen3.5-9B~\cite{qwen35blog} & 57.01 & 18.09 & 63.44 & 24.73 & 34.72 & 27.31 & 80.31 & 56.49 & 29.90 & 19.66 & 41.17 \\
\midrule
\rowcolor{gray!20}
\multicolumn{12}{c}{\textbf{Training-Specific MLLMs}}\\
\midrule
Migician~\cite{li-etal-2025-migician} & 70.64 & 45.74 & 72.76 & 67.82 & 60.07 & 72.57 & 75.76 & 84.12 & 52.58 & 33.33 & 63.54 \\
UniVG-R1~\cite{bai2025univg} & 71.97 & 58.51 & \textbf{93.13} & 76.36 & 66.32 & 81.71 & 82.83 & 88.04 & 62.89 & 44.44 & 72.64 \\
UniVG-R1$^{\dagger}$ & 76.70 & 61.70 & 87.61 & 82.18 & 64.93 & 86.28 & \textbf{87.88} & \textbf{88.25} & 69.07 & 37.61 & 74.22 \\
AdaRFT$^{\dagger}$ & 74.05 & 57.45 & 88.22 & 81.27 & 57.99 & 84.95 & 83.84 & 87.63 & 67.01 & 45.30 & 72.77 \\
\textbf{MG-Thinker} & \textbf{78.98} & \textbf{62.77} & 93.01 & \textbf{82.55} & \textbf{66.67} & \textbf{88.78} & 86.87 & \textbf{88.25} & \textbf{72.16} & \textbf{47.01} & \textbf{76.71} \\
\bottomrule
\end{tabular}
\end{adjustbox}
\label{tab1}
\end{table*}

\section{Experiments}
\subsection{Implementation Details}
\textbf{Datasets.} Stage-1 SFT uses 320K restructured MGrounding-630K samples to activate multi-image grounding capabilities. Stage-2 distills 32K MRG instances with CoT annotations, of which 25K cold-start samples couple reasoning with grounding during CoT-SFT and the remaining 7K are held out as a clean pool for BiA-DAPO's rollout groups during RL post-training. For evaluation, we assess in-domain performance on 10 MIG-Bench sub-tasks~\cite{li-etal-2025-migician} and, following UniVG-R1, further evaluate zero-shot transfer on both single-image reasoning grounding (LISA-Grounding~\cite{lai2024lisa}, LLMSeg-Grounding~\cite{wang2024llm}) and multi-frame reasoning grounding (ReVOS-Grounding~\cite{yan2024visa}, ReasonVOS-Grounding~\cite{bai2024one}) benchmarks. To probe general multi-image comprehension, we additionally test on multi-image understanding suites BLINK~\cite{fu2024blink}, MuirBench~\cite{wang2025muirbench}, MMIU~\cite{meng2024mmiu}, and MIBench~\cite{liu2024mibench}, and report standard single-image referring comprehension on RefCOCO/+/g so that gains are traced across in-domain MRG, cross-modal reasoning grounding transfer, and conventional grounding.

\noindent
\textbf{Training Details.} MG-Thinker is built on Qwen2.5-VL-7B and trained on 8$\times$A100 (80GB) with fixed random seeds for reproducibility. SFT uses learning rate $3e{-}6$ and batch size 48, while CoT-SFT and RL both use $1e{-}6$ with batch sizes 48 and 16, respectively; all stages adopt AdamW under a linear-warmup with cosine-decay schedule. During RL post-training, we sample 16 responses per query within a 1024-token cap via vLLM at temperature $1.0$, and score each rollout by the deterministic reward described in Sec.~\ref{sec:reward}. Full hyperparameters, including GIA/CRS phase schedule and clipping coefficients, are reported in Appendix~\ref{app:rl-config}.

\noindent
\textbf{Evaluation Metrics.} We adopt the standard Acc@0.5 metric following the REC protocol (IoU $\geq 0.5$), with all baselines evaluated from their officially released checkpoints under unified inference settings for fair comparison. For zero-shot reasoning grounding transfer, we keep the same Acc@0.5 protocol without any target-domain fine-tuning, so that comparisons reflect intrinsic transferability rather than task-specific adaptation, whereas multi-image understanding benchmarks retain their respective official protocols.

\begin{table}[t]
\centering
\caption{Comparison of different outputting formats. `Polling' refers to the multi-image polling approach, while `All' means outputting multi-object in one shot.}
\begin{adjustbox}{max width=\columnwidth}
\setlength{\tabcolsep}{2pt}
\renewcommand{\arraystretch}{0.85}
\small
\begin{tabular}{l|c|cccc|c}
\toprule
\textbf{Models} & \makecell{\textbf{Output}\\\textbf{Format}} & \makecell{\textbf{Common}} & \makecell{\textbf{MV}} & \makecell{\textbf{OT}} & \makecell{\textbf{Region}} & \textbf{AVG} \\
\midrule
Random Guess & / & 26.47 & 1.04 & 2.13 & 0.00 & 7.41 \\
\midrule
\multirow{3}{*}{Qwen2-VL-7B} & Poll & 19.36 & 11.81 & 20.73 & 25.95 & 19.46 \\
& All & 19.36 & 6.60 & 13.09 & 11.80 & 12.71 \\
& All+CoT & 45.71 & 9.38 & 17.55 & 15.54 & 22.05 \\
\midrule
\multirow{3}{*}{Qwen2.5-VL-7B} & Poll & 16.36 & 3.66 & 6.94 & 15.05 & 10.50 \\
& All & 27.47 & 3.90 & 15.31 & 17.29 & 15.99 \\
& All+CoT & 40.58 & 4.98 & 14.60 & 19.12 & 19.82 \\
\midrule
\multirow{3}{*}{Migician} & Poll & 72.76 & 60.07 & 67.82 & 72.57 & 68.31 \\
& All & 72.43 & 43.06 & 58.55 & 34.91 & 52.24 \\
& All+CoT & 75.56 & 41.67 & 63.82 & 41.81 & 55.72 \\
\midrule
\multirow{2}{*}{UniVG-R1} & Poll & \textbf{93.13} & 66.32 & 76.36 & 81.71 & 79.38 \\
& All & 12.99 & 30.90 & 26.55 & 35.08 & 26.38 \\
\midrule
\multirow{2}{*}{\textbf{MG-Thinker}} & Poll & 90.18 & \textbf{66.67} & \textbf{82.55} & 88.11 & 81.88 \\
& All & 93.01 & 65.22 & 81.81 & \textbf{88.78} & \textbf{82.21} \\
\bottomrule
\end{tabular}
\end{adjustbox}
\label{tab2}
\end{table}
\begin{table}[t]
\centering
\caption{Zero-shot performance on other reasoning grounding benchmarks.}
\begin{adjustbox}{max width=\columnwidth}
\setlength{\tabcolsep}{2pt}
\renewcommand{\arraystretch}{0.85}
\small
\begin{tabular}{l|ccc|cc|c}
\toprule
\multirow{2}{*}{\textbf{Models}} & \multicolumn{3}{c|}{\textbf{Single Image}} & \multicolumn{2}{c|}{\textbf{Multi Images}} & \multirow{2}{*}{\textbf{AVG}} \\
\cmidrule(lr){2-4} \cmidrule(lr){5-6}
& \textbf{LISA-val} & \textbf{LISA-test} & \textbf{LLMSeg} & \textbf{ReasVOS} & \textbf{ReVOS} & \\
\midrule
Qwen2-VL-7B & 52.00 & 49.17 & 35.53 & 9.83 & 23.55 & 34.02 \\
Qwen2.5-VL-7B & 54.06 & 50.75 & 32.48 & 11.14 & 26.81 & 35.04 \\
Migician & 36.00 & 32.09 & 34.68 & 33.41 & 39.70 & 35.18 \\
UniVG-R1 & 64.00 & 59.69 & \textbf{50.60} & 58.73 & 60.03 & 58.61 \\
\textbf{MG-Thinker} & \textbf{64.29} & \textbf{60.10} & 49.94 & \textbf{59.47} & \textbf{61.32} & \textbf{59.02} \\
\bottomrule
\end{tabular}
\end{adjustbox}
\label{tab3}
\end{table}
\subsection{Main Experimental Results}
\textbf{Analysis of Reasoning Paradigms.}
\looseness=-1
We compare three reasoning paradigms for MRG: (i) direct end-to-end prediction without explicit reasoning, (ii) prompt-induced CoT, and (iii) MG-Thinker at cold-start vs.\ fully trained stages. Across representative sub-tasks, direct prediction often fails to establish reliable cross-image correspondences, while CoT prompting may suffer from reasoning drift that misleads the eventual localization. MG-Thinker instead shows progressively improved reasoning--grounding consistency: the cold-start model can localize correctly but produces unstable, sometimes drifting rationales, whereas the fully trained model consistently follows a reliable ``anchor-reason-ground'' pattern that yields faithful thinking and accurate predictions. Detailed case studies and qualitative visualizations are provided in \mbox{Appendix~\ref{app:reasoning-cases}}.

\noindent
\looseness=-1
\textbf{Performance on MIG-Bench.} Tab.~\ref{tab1} compares MG-Thinker with state-of-the-art MLLMs (Qwen2/2.5/3-VL/3.5~\cite{bai2025qwen3,qwen3.5}, Mantis, LLaVA-OV, MiniCPM-V-2.6, mPLUG-Owl3, InternVL2/3) and two specialized baselines: Migician, the first end-to-end multi-image grounding paradigm that bridges multi-image understanding with fine-grained single-image perception, and UniVG-R1, a recent GRPO-style attempt that couples multi-image localization with grounding-specific loss augmentations. To further isolate our bi-axial design at the algorithm level, we faithfully re-implement two single-axis baselines on a shared GRPO backbone under our CoT-SFT initialization and 25K MRG data: AdaRFT$^{\dagger}$ (offline difficulty curriculum, competence axis only) and UniVG-R1$^{\dagger}$ (mIoU-weighting, signal axis only). UniVG-R1$^{\dagger}$ already surpasses the original UniVG-R1 (74.22 vs.\ 72.64), validating the effectiveness of our curated data. Nonetheless, neither single-axis variant catches MG-Thinker, which sets a new state of the art on MIG-Bench on average (76.71 vs.\ 74.22 for the strongest baseline), beating Migician by 13.2 points and UniVG-R1 by 4.1 points and confirming that only the bi-axial treatment of both signal and competence axes yields the full gain. Despite using only 7B parameters, MG-Thinker outperforms 70B-scale models such as Qwen2.5-VL-72B by 30.5 points on average, highlighting a strong performance/efficiency trade-off. Detailed qualitative visual comparisons and case-study analysis are deferred to \mbox{Appendix~\ref{app:qualitative}}.
\begin{table}[t]
\centering
\caption{Cross-task heterogeneity analysis on MIG-Bench. Tasks are grouped into V-Het, R-Het, and L-Het.}
\label{tab4}
\begin{adjustbox}{max width=\columnwidth}
\setlength{\tabcolsep}{2pt}
\renewcommand{\arraystretch}{0.85}
\small
\begin{tabular}{l|ccc|cc}
\toprule
\textbf{Task group} & \textbf{BiA-DAPO} & \textbf{UniVG-R1$^{\dagger}$} & \textbf{AdaRFT$^{\dagger}$} & \textbf{$\Delta$Uni$^{\dagger}$} & \textbf{$\Delta$Ada$^{\dagger}$} \\
\midrule
V-Het (mean-drift)     & \textbf{76.14} & 75.69 & 71.73 & +0.45 & \textbf{+4.41} \\
R-Het (variance-drift) & \textbf{67.24} & 62.95 & 64.53 & \textbf{+4.29} & +2.71 \\
L-Het (no dominant)    & \textbf{85.76} & 83.53 & 82.41 & +2.23 & +3.35 \\
\bottomrule
\end{tabular}
\end{adjustbox}
\end{table}
\begin{table}
\centering
\caption{Performance comparison on multi-image understanding benchmarks.}
\label{tab5}
\begin{adjustbox}{max width=\columnwidth}
\setlength{\tabcolsep}{2pt}
\renewcommand{\arraystretch}{0.85}
\small
\begin{tabular}{l|ccccc}
\toprule
\textbf{Models} & \textbf{MuirBench} & \textbf{BLINK Val} & \textbf{MIBench} & \textbf{MMIU} & \textbf{AVG} \\
\midrule
LLaVA-1.5 & 23.46 & 37.13 & 26.83 & 19.20 & 26.66 \\
CogVLM & 20.85 & 41.54 & ,  & 23.57 & 28.65 \\
Idefics2-8B & 26.08 & ,  & 46.39 & 27.80 & 33.42 \\
mPLUG-Owl3 & 39.67 & 50.30 & 56.66 & 21.72 & 42.09 \\
InternVL2-8B & 48.70 & 50.57 & 52.91 & 42.00 & 48.55 \\
Mantis & 44.50 & 49.05 & 45.09 & 45.60 & 46.06 \\
LLaVA-OV-7B & 41.80 & 48.20 & 71.29 & 44.46 & 51.44 \\
MiniCPM-V 2.6 & 42.65 & 51.45 & 71.09 & 50.19 & 53.85 \\
Qwen2-VL-7B & 39.88 & 52.35 & 68.06 & 54.36 & 53.66 \\
Migician & 57.81 & 51.53 & 71.42 & \textbf{54.65} & 58.85 \\
UniVG-R1 & 44.61 & 51.77 & 66.93 & 53.13 & 54.11 \\
\midrule
\textbf{MG-Thinker} & \textbf{61.77} & \textbf{54.13} & \textbf{71.88} & 53.93 & \textbf{60.43} \\
\bottomrule
\end{tabular}
\end{adjustbox}
\end{table}

\begin{table*}[t]
\centering
\caption{Ablation study of different training stages and RL post-training algorithm components.}
\label{tab6}
\begin{adjustbox}{max width=\linewidth}
\setlength{\tabcolsep}{4pt}
\renewcommand{\arraystretch}{0.85}
\small
\begin{tabular}{l|c|cc|c|cccc|c|cc|c}
\toprule
\multirow{3}{*}{\textbf{Methods}} & \multirow{3}{*}{\textbf{No.}} &
\multicolumn{3}{c|}{\textbf{Spontaneous Grounding}} &
\multicolumn{7}{c|}{\textbf{Referential Grounding}} &
\multirow{3}{*}{\textbf{AVG}} \\
\cmidrule(lr){3-5} \cmidrule(lr){6-12}
& &
\multicolumn{2}{c|}{\textbf{Difference}} & \textbf{Similarity} &
\multicolumn{4}{c|}{\textbf{Visual Reference}} &
\textbf{Textual} &
\multicolumn{2}{c|}{\textbf{Visual+Textual}} & \\
\cmidrule(lr){3-4} \cmidrule(lr){5-5}
\cmidrule(lr){6-9} \cmidrule(lr){10-10} \cmidrule(lr){11-12}
& &
\textbf{Static} & \textbf{Robust} & \textbf{Common} &
\textbf{OT} & \textbf{MV} & \textbf{Region} & \textbf{Refer} &
\textbf{GG} &
\textbf{Reason} & \textbf{Co-Re} & \\
\midrule
\rowcolor{gray!20}
\multicolumn{13}{c}{\textbf{Baseline MLLMs}}\\
\midrule
Qwen2-VL-7B~\cite{wang2024qwen2}      & 1 & 27.84 & 38.30 & 19.36 & 20.73 & 11.81 & 25.95 & 23.23 & 58.52 & 48.51 & 11.97 & 28.62 \\
Qwen2.5-VL-7B~\cite{bai2025qwen2}    & 2 & 29.92 & 22.45 & 16.36 &  6.94 &  3.66 & 15.05 & 35.67 & 63.51 &  4.03 &  3.42 & 20.10 \\
Qwen3-VL-8B~\cite{bai2025qwen3} & 3 & 29.36 & 24.47 & 58.83 & 28.00 & 17.36 & 23.44 & 33.33 & 70.93 & 32.99 & 10.25 & 32.90 \\
Qwen3.5-9B~\cite{qwen35blog} & 4 & 57.01 & 18.09 & 63.44 & 24.73 & 34.72 & 27.31 & 80.31 & 56.49 & 29.90 & 19.66 & 41.17 \\
\midrule
\rowcolor{gray!20}
\multicolumn{13}{c}{\textbf{Specific Training Stages}}\\
\midrule
SFT             & 5 & 65.49 & 52.13 & 86.19 & 75.68 & 61.81 & 77.89 & 82.83 & 83.85 & 57.73 & 35.90 & 67.95 \\
CoT-SFT         & 6 & 72.18 & 60.64 & 85.72 & 74.89 & 63.54 & 77.47 & 81.82 & 85.90 & 69.07 & 41.03 & 71.23 \\
\midrule
\rowcolor{gray!20}
\multicolumn{13}{c}{\textbf{Specific Algorithm Components}}\\
\midrule
GRPO            & 7 & 70.27 & 60.83 & 87.50 & 80.00 & 65.26 & 82.49 & 82.44 & 87.63 & 72.65 & 44.30 & 73.34 \\
DAPO            & 8 & 72.55 & 60.97 & 85.15 & 75.27 & 60.76 & 87.03 & 84.23 & 87.42 & 73.20 & 43.22 & 72.98 \\
w/o. GIA         & 9 & 78.79 & 60.64 & 90.80 & 78.91 & \textbf{67.01} & 86.70 & 85.86 & \textbf{89.69} & \textbf{76.29} & 41.03 & 75.57 \\
w/o. CRS         & 10 & 77.46 & 58.51 & 90.92 & \textbf{83.45} & 62.15 & 88.04 & 86.53 & 85.86 & 75.26 & 40.17 & 74.84 \\
\textbf{MG-Thinker} & 11 & \textbf{78.98} & \textbf{62.77} & \textbf{93.01} & 82.55 & 66.67 & \textbf{88.78} & \textbf{86.87} & 88.25 & 72.16 & \textbf{47.01} & \textbf{76.71} \\
\bottomrule
\end{tabular}
\end{adjustbox}
\end{table*}
\begin{figure*}[h]
    \begin{minipage}[t]{0.49\textwidth}
        \centering
        \includegraphics[width=0.85\linewidth]{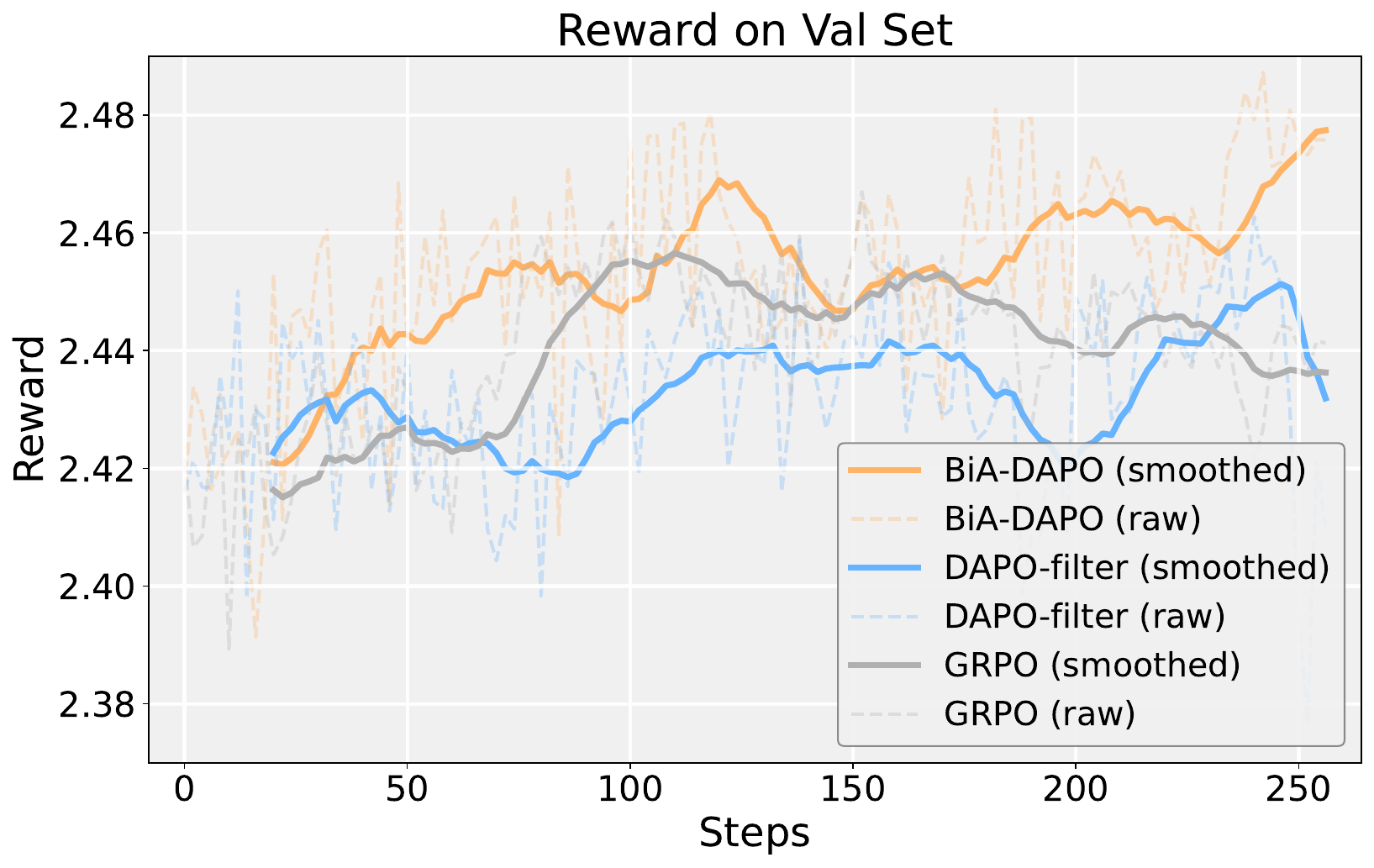}
        \caption{PFA reward dynamics under identical training. BiA-DAPO sustains a smooth upward trajectory, whereas GRPO/DAPO oscillate non-monotonically due to coupled sparsity and misalignment.}
        \label{fig:pfa-dynamics}
    \end{minipage}
    \hfill
    \begin{minipage}[t]{0.49\textwidth}
        \centering
        \includegraphics[width=0.8\linewidth]{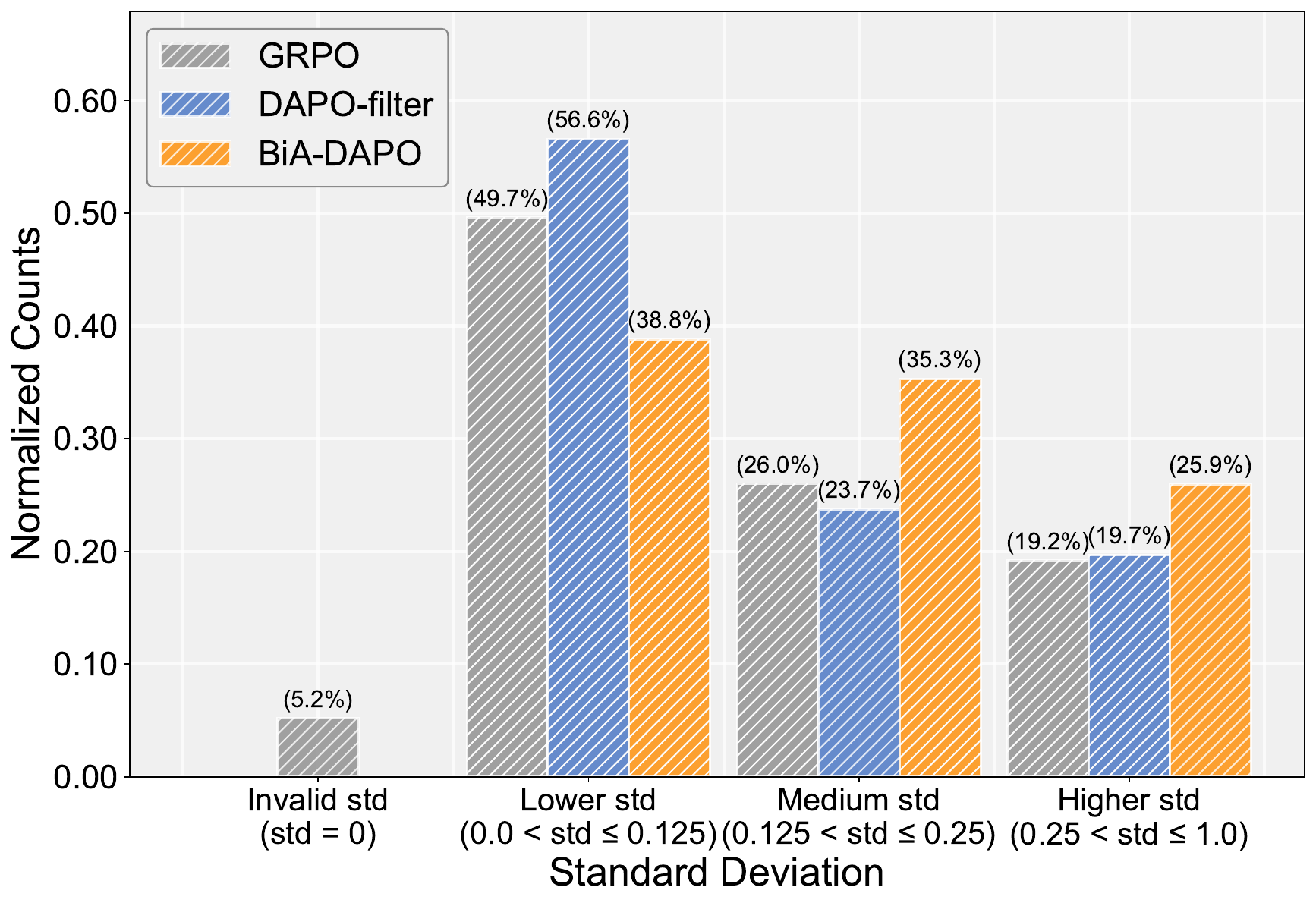}
        \caption{Per-step distribution of group-level $\sigma_g$. BiA-DAPO concentrates on medium/high informativeness intervals (GIA gating), while GRPO/DAPO over-retain low-$\sigma_g$ groups whose advantage is mathematically zero.}
        \label{fig:sigma-dist}
    \end{minipage}
\end{figure*}
\noindent
\textbf{Effectiveness of reinforcement post-training.}
\looseness=-1
Fig.~\ref{fig:pfa-dynamics} compares BiA-DAPO's reward dynamics with GRPO/DAPO under identical training settings: BiA-DAPO sustains a smooth upward trajectory with markedly smaller step-to-step fluctuation, whereas GRPO/DAPO oscillate without consistent improvement and even intermittently regress, reflecting their inability to disentangle the two MRG pathologies. Fig.~\ref{fig:sigma-dist} reports the per-step distribution of group-level $\sigma_g$ (binned at $0$, $(0,0.125]$, $(0.125,0.25]$, $(0.25,1.0]$), corresponding to advantage-vanished, low-, medium-, and high-informativeness groups: BiA-DAPO concentrates on the medium/high intervals (GIA retains advantage-bearing groups), while DAPO skews toward the low end and GRPO further over-retains $\sigma_g{=}0$ groups whose gradient is mathematically zero, the drifting sparsity that drives GRPO/DAPO's reward oscillation. Together, the smooth trajectory in Fig.~\ref{fig:pfa-dynamics} and the healthier $\sigma_g$ mass in Fig.~\ref{fig:sigma-dist} give direct evidence that GIA and CRS act as an axis-decoupled treatment of the two pathologies rather than merely a single stabilization knob; two complementary diagnostics in \mbox{Appendix~\ref{app:dynamics}} further substantiate this joint GIA-CRS mechanism.

\noindent
\textbf{Cross-task heterogeneity analysis.} We partition all MRG sub-tasks into three groups by intrinsic heterogeneity: \emph{Visual-}, \emph{Reasoning-}, and \emph{Low-Heterogeneity}. V-Het tasks exhibit mean-reward drift (corresponding to the competence axis), R-Het tasks exhibit variance drift (corresponding to the signal axis), while L-Het tasks show neither prominently (group-wise task composition in Appendix~\ref{app:mig-tasks}). Tab.~\ref{tab4} then reveals complementary blind spots: UniVG-R1$^{\dagger}$ matches BiA-DAPO on V-Het ($+0.45$) but lags on R-Het ($+4.29$), whereas AdaRFT$^{\dagger}$ stays close on R-Het ($+2.71$) but loses on V-Het ($+4.41$). Only BiA-DAPO leads on both axes simultaneously, confirming that GIA and CRS act on genuinely complementary pathologies.

\noindent
\textbf{One-shot multi-object grounding.} Beyond single-object, MG-Thinker performs one-shot multi-object localization across images in a single inference, challenging for both reasoning-based (UniVG-R1) and non-reasoning (Migician) baselines. Tab.~\ref{tab2} compares polling, single-query multi-object, and CoT-prompted modes: MG-Thinker's one-shot output matches its polling accuracy on average (82.21 vs.\ 81.88) and exceeds it on Common and Region, while others degrade markedly in the one-shot setting, indicating our approach also mitigates adverse post-training effects. This suggests the CoT-driven reasoning trace plans a joint cross-image layout before localization, rather than degenerating into per-image polling like the baselines.

\noindent
\textbf{Zero-Shot on Reasoning Grounding Benchmarks.} Tab.~\ref{tab3} reports zero-shot results on single-image (LISA/LLMSeg-Grounding) and video (ReVOS/ReasonVOS-Grounding) reasoning grounding benchmarks. MG-Thinker substantially surpasses Migician and matches UniVG-R1 (59.02\% avg.), the slight LLMSeg-Grounding drop reflects UniVG-R1's use of single-image reasoning data during RL, which we omit. These results show transferability without task-specific tuning. Notably, the multi-frame gains (ReVOS/ReasonVOS) emerge purely from MG-Thinker's cross-image reasoning priors without any video-specific supervision, showing that multi-image reasoning transfers naturally to the temporal setting.

\noindent
\textbf{Multi-image Understanding.} Tab.~\ref{tab5} shows MG-Thinker delivers strong performance across diverse multi-image understanding suites. While UniVG-R1's grounding gains transfer poorly (falling behind Migician on several benchmarks), MG-Thinker yields superior overall results and notably leads on MuirBench counting/visual-grounding subtasks, indicating BiA-DAPO benefits generalized understanding rather than task-specific gains. This further suggests that MG-Thinker's grounding-anchored CoT strengthens the cross-image attention that also underpins general multi-image comprehension, avoiding over-specialization to the grounding objective.

\noindent
\textbf{General single-image grounding.} On standard RefCOCO/+/g (full table is in Appendix~\ref{app:refcoco}), MG-Thinker remains strongly competitive on single-image referring expression grounding, indicating stable multi-modal generalization beyond reasoning-centric benchmarks. This further confirms that our MRG-oriented post-training preserves the base-model grounding prior rather than over-specializing to multi-image tasks, a common over-specialization trap of task-adaptive RL. We attribute this stability to BiA-DAPO's group-relative updates being anchored on rollout-level reward statistics rather than any modality-specific loss term, so the RL signal never explicitly suppresses the single-image grounding prior even when the training pool is dominated by multi-image samples.

\subsection{Ablation Studies}
We validate MG-Thinker from two perspectives: training stages and BiA-DAPO components.

\noindent
\textbf{Effect of Training Stages.} Tab.~\ref{tab6} validates the three progressive stages, grounding activation, reasoning--grounding integration, and reasoning enhancement. Stage-1 SFT alone surpasses Qwen2/2.5-VL-7B across diverse sub-tasks including visual comparison, object tracking, and spatial perception, achieving over 2$\times$ average-accuracy gain on MIG-Bench. Stage-2 CoT-SFT (cold-start) then couples grounding with explicit reasoning, yielding notable improvements on correspondence- and reasoning-intensive sub-tasks. Finally, BiA-DAPO delivers the optimum on top of CoT-SFT, together confirming the progressive realization of the proposed MRG paradigm from activate to enhance.

\noindent
\textbf{Effect of BiA-DAPO Components.} We perform controlled ablations on the RL algorithm (Tab.~\ref{tab6}). GRPO and DAPO baselines yield improvements but are upper-bounded by their single-statistic treatment (Sec.~\ref{sec:biadapo}), which conflates advantage magnitude and competence position into one number. Removing either axis-specific mechanism causes complementary degradation: \textit{w/o.\ GIA} (Axis-1 disabled) loses on tasks with wide intra-group spread (OT, Co-Re), where informativeness gating is critical for relieving drifting advantage-signal sparsity, since the losses trace back to $\sigma_g{\approx}0$ groups that unfiltered DAPO would otherwise admit; \textit{w/o.\ CRS} (Axis-2 disabled) loses on tasks with wide between-group competence spread (Robust, MV, GG), where ordered $\bar{r}_g$ traversal is needed to remedy competence-difficulty misalignment, since these are precisely the sub-tasks whose group means shift the most across training. Integrating both axes recovers the optimum, corroborating the cross-heterogeneity evidence in Tab.~\ref{tab4}. Detailed sensitivity ablations on the GIA phase schedule and CRS band width are reported in Appendix~\ref{app:sensitivity}.

\section{Conclusion}
In this paper, we present MG-Thinker, an RL-based framework for multi-image reasoning grounding (MRG). We identify two MRG-specific challenges largely overlooked by prior work: an inherent coarse-to-fine hierarchical pattern from semantic reasoning to region-level localization, and a pronounced task-sample difficulty heterogeneity that induces drifting advantage-signal sparsity and competence-difficulty misalignment during post-training. To address them, we curate a 25K MRG-specific CoT dataset that instills the hierarchical pattern, and propose Bi-Axial DAPO (BiA-DAPO), which disentangles rollout advantages along an intra-group signal axis and an inter-group competence axis via Group-Informativeness Assessment and Cascaded-Reward Stratification over a shared candidate pool. Extensive experiments confirm state-of-the-art MRG performance with consistent generalization to broader multimodal benchmarks, and we hope this bi-axial view of group-relative RL offers a broadly useful design pattern for other hierarchical, difficulty-heterogeneous multimodal reasoning tasks.

\bibliographystyle{ACM-Reference-Format}
\bibliography{sample-base}

\clearpage
\appendix
\twocolumn[
\begin{center}
{\LARGE\bfseries MG-Thinker: Bi-Axial Self-Reflection for Multi-Image \\ Reasoning Grounding\par}
\vspace{0.6em}
{\large Appendix\par}
\end{center}
\vspace{1.0em}
]
\section{Multi-Image Grounding Tasks Categorization}
\label{app:mig-tasks}

\noindent\textbf{Original taxonomy of multi-image grounding tasks.} Prior multi-image grounding work~\cite{li-etal-2025-migician} formulates ten sub-tasks under two top-level branches, distinguished by whether the query carries an explicit referential cue. \emph{Spontaneous Grounding} (SG) requires the model to autonomously discover and localize a target through cross-image relations, comprising Static Difference Grounding (Static), Robust Difference Grounding (Robust), and Common Object Grounding (Common), the difficulty here lies in identifying \emph{what to localize} purely from cross-image relations. \emph{Referential Grounding} (RG) provides an explicit cue and is further split by cue modality into textual cue (Group Grounding, GG), visual cue (Object Tracking, OT; Multi-View Grounding, MV; Visual Referring Grounding, Refer; Region Locating, Region), and combined visual-textual cue (Reasoning Grounding, Reason; Correspondence, Co-Re); the difficulty here lies in correctly interpreting the cue and propagating it across images. All sub-tasks are evaluated with the standard $\text{Acc}_{0.5}$ IoU criterion, and we adopt these abbreviations whenever a sub-task is referenced throughout the paper.

\begin{figure*}[!t]
  \centering
  \includegraphics[width=0.9\linewidth]{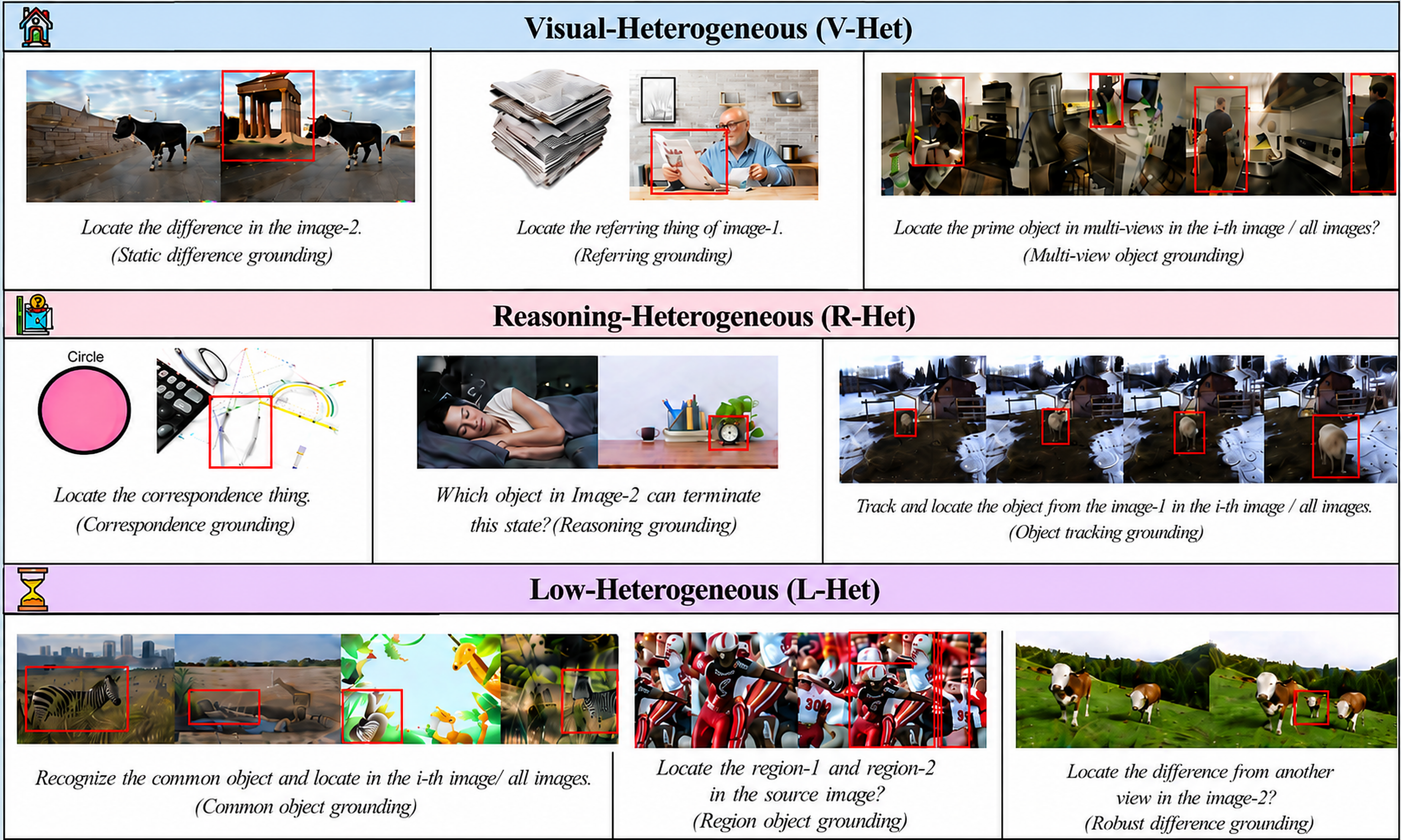}
  \caption{Restructured multi-image grounding data, organized by intrinsic task heterogeneity into Visual-Heterogeneity (V-Het), Reasoning-Heterogeneity (R-Het), and Low-Heterogeneity (L-Het) groups. Each group spans both single-object and one-shot multi-object grounding, reflecting the distinct challenges of MRG.}
  \label{fig:app-mig-tasks}
\end{figure*}

\noindent\textbf{Intrinsic heterogeneity-based re-categorization.} The cue-based partition above, while convenient for benchmarking, does not directly reveal the MRG-specific source of difficulty during post-training. We therefore re-organize the same ten sub-tasks along two largely orthogonal axes of intrinsic heterogeneity, \emph{cross-image visual variation} (e.g., viewpoint shifts, repeated or distractor instances, layout changes that alter how the same scene or entity appears across images) and \emph{cross-image reasoning depth} (e.g., temporal tracking, abstract correspondence, referential semantic logic that requires multi-step inference over the image set). The two axes manifest at the rollout-statistics level as the inter-group competence axis (mean reward $\bar{r}_g$) and the intra-group signal axis (variance $\sigma_g^2$), respectively, on which the cross-task heterogeneity analysis of Tab.~\ref{tab4} (main paper) is based. Different sub-tasks load these two axes very differently, and we group them by the dominant source of difficulty into three intrinsic categories, as shown in Fig.~\ref{fig:app-mig-tasks}:
\begin{itemize}
    \item \emph{Visual-Heterogeneity} (V-Het): tasks where intrinsic heterogeneity concentrates on the visual axis, manifesting as pronounced mean-reward drift across sub-tasks (the inter-group competence axis); e.g., multi-view, group, and referential grounding, whose challenge lies in reconciling appearance variations of the same target across images.
    \item \emph{Reasoning-Heterogeneity} (R-Het): tasks where intrinsic heterogeneity concentrates on the reasoning axis, manifesting as pronounced variance drift within rollout groups (the intra-group signal axis); e.g., object tracking, correspondence, and reasoning grounding, whose difficulty escalates with the depth of cross-image inference required.
    \item \emph{Low-Heterogeneity} (L-Het): tasks where neither axis dominates, with both mean and variance remaining stable; e.g., common-object localization in static scenes, exhibiting relatively uniform per-sample difficulty.
\end{itemize}

Each group spans both single-object and one-shot multi-object grounding, reflecting the distinct challenges of MRG. The Cross-task heterogeneity analysis in Tab.~\ref{tab4} (main paper) groups sub-tasks accordingly, isolating the complementary effects of GIA (signal axis) and CRS (competence axis) on each group.

\section{CoT Prompt Templates for MRG-Specific Activation}
\label{app:templates}
Unlike prior CoT activation that targets single-image reasoning or generic visual QA, our prompt templates are purpose-built around two MRG-specific characteristics: (i) the joint presence of \emph{visual} and \emph{textual} referential cues that must be cross-interpreted across multiple images, and (ii) a \emph{coarse-to-fine hierarchy} that runs through both reasoning and grounding (CoT$\to$image-id$\to$bbox), so that semantic-level inference and spatial-level localization unfold along the same trajectory. Crucially, the templates also explicitly support \emph{one-shot multi-image multi-target} localization, an output mode rarely formulated by existing reasoning-grounding pipelines. We organize templates along the four MRG-specific task types categorized in §3.2, \emph{visual comparative analysis}, \emph{spatial perception}, \emph{temporal perception}, and \emph{visual semantic/logical association}, and use the stronger Qwen2.5-VL-72B annotator to instantiate them, so that multi-image evidence and multi-perspective thinking jointly guide reasoning and reasoning, in turn, sharpens grounding precision.

\begin{figure*}[!t]
  \centering
  \includegraphics[width=0.95\linewidth]{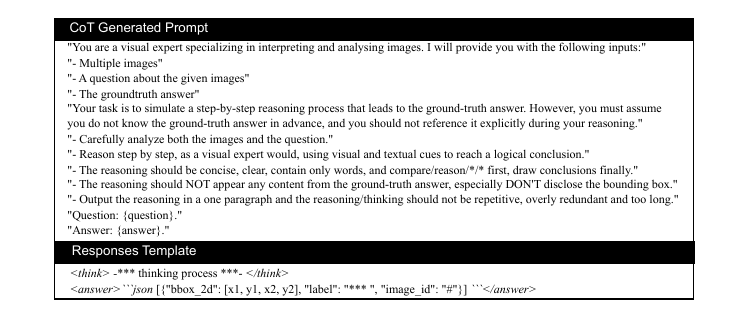}
  \caption{CoT templates designed for the four MRG-specific task types and the standard response format. $*$ denotes textual content and $\#$ denotes natural numbers; in the generated CoT templates, $*$ is replaced with task-specific descriptive terms.}
  \label{fig:app-templates}
\end{figure*}

As illustrated in Fig.~\ref{fig:app-templates}, each template comprises:
\begin{itemize}
    \item \textbf{Role definition.} The model is positioned as a visual expert specialized in cross-image interpretation and analysis.
    \item \textbf{Input specification.} Multiple images, the user query, and the ground-truth answer are provided; the latter is used solely as a verification anchor and is never disclosed during the thinking phase.
    \item \textbf{Task-oriented operators.} Per-type instructions inject \emph{comparison}, \emph{searching}, \emph{observation}, \emph{tracking}, and \emph{association} cues, encouraging step-wise multi-perspective evidence-seeking that integrates visual and textual referential cues.
    \item \textbf{Reasoning constraints.} Explicit prohibitions block leakage of ground-truth answers or bounding-box coordinates inside the \textless think\textgreater{} block, ensuring genuinely explanatory rationales rather than answer-leaking shortcuts.
    \item \textbf{Hierarchical output format.} A structured \textless think\textgreater{}\textless /think\textgreater{} trace, naturally partitioned into a coarse-to-fine semantic hierarchy, is followed by a JSON-formatted \textless answer\textgreater{}\textless /answer\textgreater{} block carrying multi-image bounding-box coordinates with explicit \texttt{image\_id} keys, enabling one-shot multi-image multi-target grounding within a single inference.
\end{itemize}

The two-step IoU-improvement post-processing (Sec.~3.3) then filters pseudo-reasoning samples and retains only rationales that demonstrably improve grounding, yielding the final 25K cold-start corpus.

\section{DAPO Clipped Surrogate Objective and Group-Relative Advantage}
\label{app:dapo}
For completeness we recap the DAPO clipped surrogate on top of which BiA-DAPO is built. Given a question--answer pair $(q,a)$ and a sampled rollout group $\{o_i\}_{i=1}^{G}$ drawn from the old policy $\pi_{\theta_{\text{old}}}$, DAPO optimizes
\begin{multline}
\mathcal{J}_{\text{DAPO}}(\theta) = \mathbb{E}_{(q,a),\,\{o_i\}\sim \pi_{\theta_{\text{old}}}}\!\Big[ \tfrac{1}{\sum_i|o_i|}\!\sum_{i,t} \\
\min\!\big( r_{i,t}\hat{A}_{i,t},\,\mathrm{clip}(r_{i,t},\,1{-}\varepsilon_l,\,1{+}\varepsilon_h)\hat{A}_{i,t} \big) \Big],
\end{multline}
with the per-token importance ratio
\begin{equation}
r_{i,t}(\theta) = \frac{\pi_{\theta}(o_{i,t}\mid q,o_{i,<t})}{\pi_{\theta_{\text{old}}}(o_{i,t}\mid q,o_{i,<t})},
\end{equation}
and asymmetric clip-higher coefficients $\varepsilon_l{=}0.2$, $\varepsilon_h{=}0.28$. The group-relative advantage of rollout $i$ is
\begin{equation}
\hat{A}_i \;=\; \frac{\mathrm{acc}_i \,-\, \bar{r}_g}{\sigma_g + \varepsilon},
\end{equation}
where $\bar{r}_g$ and $\sigma_g$ are the within-group mean and standard deviation of $\mathrm{acc}_i{=}R_{\text{id},i}{+}R_{\text{IoU},i}$, and $\varepsilon{=}10^{-6}$ is a numerical guard. BiA-DAPO retains this surrogate form and additionally reuses the orthogonal pair $(\bar{r}_g,\sigma_g)$ as Axis-2/Axis-1 statistics for sample-level intervention by GIA and CRS (Sec.~\ref{sec:biadapo}).

\begin{figure*}[t]
    \begin{minipage}[t]{0.48\textwidth}
        \centering
        \includegraphics[width=0.8\linewidth]{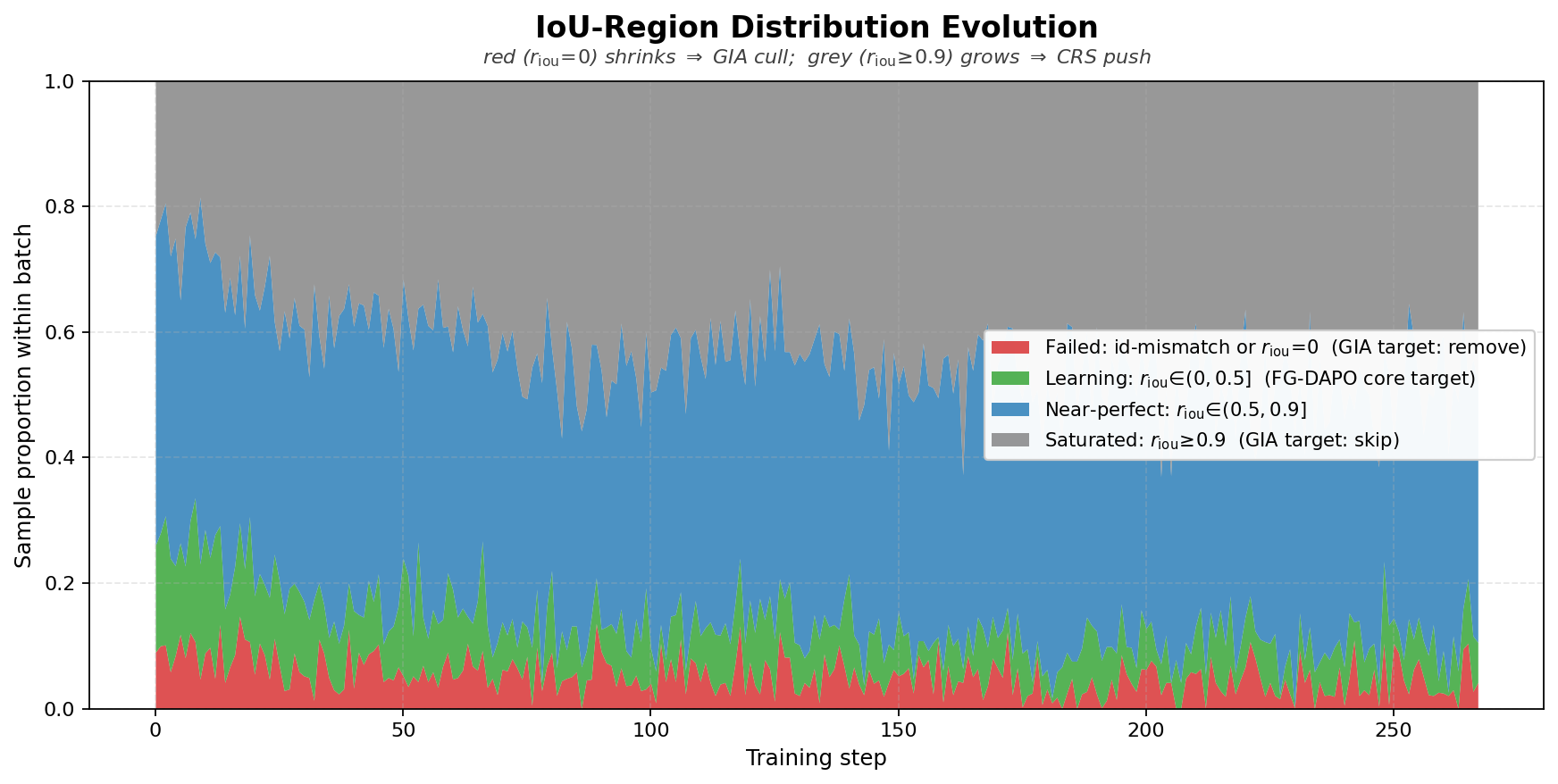}
        \caption{Per-step $R_{\text{IoU}}$ outcome shares (\emph{failed}/\emph{partial}/\emph{saturated}) across training. BiA-DAPO progressively shrinks the \emph{failed} share (GIA) and grows the \emph{saturated} share (CRS), whereas GRPO/DAPO retain a large failed share due to drifting advantage-signal sparsity.}
        \label{fig:app-iou-dynamics}
    \end{minipage}
    \hfill
    \begin{minipage}[t]{0.5\textwidth}
        \centering
        \includegraphics[width=1.0\linewidth]{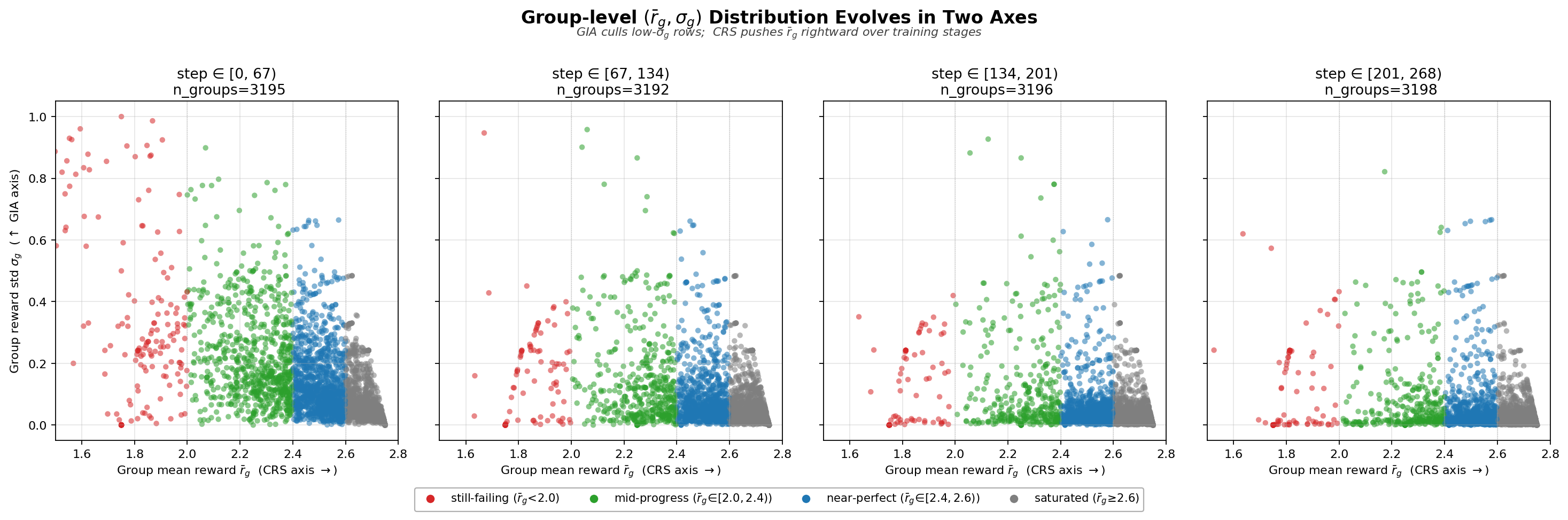}
        \caption{Bi-axial $(\bar{r}_g, \sigma_g)$ density migration during training. BiA-DAPO migrates mass diagonally from upper-left (low competence, high spread) to lower-right (high competence, moderate spread), corroborating the joint Axis-1$+$Axis-2 intervention by GIA and CRS.}
        \label{fig:app-axis-dynamics}
    \end{minipage}
\end{figure*}
\begin{table*}[t]
\centering
\caption{RL configuration in VeRL for reinforcement post-training. Basic settings (learning rate, hardware, global schedule) are specified in §4.1.}
\label{tab:app-rl-settings}
\setlength{\tabcolsep}{6pt}
\begin{adjustbox}{max width=\textwidth}
\setlength{\tabcolsep}{4pt}
\begin{tabular}{@{} p{3.0cm} | p{4.5cm} | p{8.5cm} | p{3.0cm} @{}}
\toprule
\textbf{Setting} & \textbf{Value} & \textbf{Rationale} & \textbf{Where used} \\
\midrule
Training batch / Candidate pool
  & 16 / 48
  & Pool follows DAPO's ``generation batch'' ($3{\times}$ training batch) for stable axis estimation
  & Rollout / selection \\[2pt]
Max response length
  & 1024
  & Matches the RL post-training generation budget across all methods
  & Rollout generation \\[2pt]
Responses per prompt
  & 16
  & Fixed group size for within-group statistics (GIA) and reward-mean ranking (CRS)
  & Rollout sampling \\[2pt]
Sampling (train)
  & temp\,=\,1.0, top-$p$\,=\,1.0, top-$k$\,=\,$-$1
  & Default exploration-oriented sampling
  & Rollout generation \\[2pt]
Sampling (val)
  & top-$p$\,=\,0.7, do\_sample\,=\,true, $n$\,=\,1
  & Conservative evaluation sampling
  & Validation rollout \\[2pt]
PPO clipping
  & $\varepsilon_l$\,=\,0.2, $\varepsilon_h$\,=\,0.28
  & Asymmetric clip-higher in our VeRL configuration
  & Policy optimization \\[2pt]
Loss aggregation
  & token-mean
  & Normalization for variable-length sequences
  & Policy optimization \\[2pt]
KL usage
  & use\_kl\_in\_reward\,=\,false; use\_kl\_loss\,=\,true
  & KL regularization applied as a loss term
  & Regularization \\[2pt]
KL coefficients
  & kl\_loss\_coef\,=\,0.01; kl\_coef\,=\,0.001
  & Fixed coefficients in our RL setting
  & Regularization \\[2pt]
KL type
  & low\_var\_kl
  & Low-variance KL formulation in VeRL
  & Regularization \\
\bottomrule
\end{tabular}%
\end{adjustbox}
\end{table*}
\section{Empirical Dynamics of GIA and CRS}
\label{app:dynamics}
We present two complementary diagnostic views that empirically substantiate the bi-axial mechanism of BiA-DAPO. The two views correspond, respectively, to within-group (Axis-1) and joint Axis-1$\times$Axis-2 interventions discussed in Sec.~\ref{sec:biadapo}, and underpin the qualitative claims in §4.2 \emph{Effectiveness of reinforcement post-training}.

\noindent\textbf{Per-step $R_{\text{IoU}}$ outcomes (Axis-1 view).} Fig.~\ref{fig:app-iou-dynamics} stratifies per-step rollouts into three IoU regions: \emph{failed} ($R_{\text{IoU}}{=}0$), \emph{partial} (linearly interpolated), and \emph{saturated} ($R_{\text{IoU}}{=}1$). Under GRPO/DAPO, the \emph{failed} share remains persistently large, manifesting as $\sigma_g{\approx}0$ collapsed groups and confirming drifting advantage-signal sparsity along Axis-1. BiA-DAPO progressively shrinks the \emph{failed} share, GIA's variance-driven gating retains advantage-bearing groups, and grows the \emph{saturated} share, CRS's competence cascade pulls the policy into higher-competence strata, evidencing that the two pathologies are jointly relieved.

\noindent\textbf{Bi-axial $(\bar{r}_g, \sigma_g)$ density migration (Axis-1$\times$Axis-2 view).} Fig.~\ref{fig:app-axis-dynamics} visualizes the joint distribution of group-level $(\bar{r}_g, \sigma_g)$ across training steps. GRPO/DAPO accumulate mass in the upper-left quadrant (low competence, high spread) and the lower-left strip (low competence, near-zero spread), reflecting both misalignment and sparsity. BiA-DAPO migrates the density diagonally toward the lower-right quadrant (high competence, moderate spread), precisely the trajectory predicted by GIA$+$CRS coupling, where Axis-1 sparsity is absorbed by GIA gating while Axis-2 misalignment is corrected by CRS stratification.

Together, these two views provide direct evidence that GIA and CRS act on \emph{complementary} statistical axes rather than redundant ones, consistent with the cross-task heterogeneity analysis in Tab.~\ref{tab4}.
\begin{table*}[!t]
\centering
\caption{Single-axis corner-case ablations of BiA-DAPO on MIG-Bench. \textbf{Zero-Threshold}: $\alpha_0{=}0$ with pool $48$ (open gate, equivalent to \emph{w/o.\ GIA}). \textbf{Unexpanded Pool}: pool $=16$ (no candidate-pool expansion). \textbf{Static-$\alpha$}: $\alpha{=}0.05$ frozen (no phase decay). All variant rows use the Poll output format; the MG-Thinker row reports the better of Poll/All on the four multi-target sub-tasks, as in Tab.~\ref{tab1}.}
\label{tab:app-sensitivity-ablation}
\begin{adjustbox}{max width=\textwidth}
\setlength{\tabcolsep}{4pt}
\renewcommand{\arraystretch}{1.0}
\small
\begin{tabular}{l|cccccccccc|c}
\toprule
\textbf{Setting} & \textbf{Static} & \textbf{Robust} & \textbf{Common} & \textbf{OT} & \textbf{MV} & \textbf{Region} & \textbf{Refer} & \textbf{GG} & \textbf{Reason} & \textbf{Co-Re} & \textbf{AVG} \\
\midrule
Zero-Threshold     & 78.79 & 60.64 & 90.80 & 78.91 & 67.01 & 86.70 & 85.86 & \textbf{89.69} & \textbf{76.29} & 41.03 & 75.57 \\
Unexpanded Pool    & 77.43 & 60.64 & 91.17 & 81.81 & \textbf{67.01} & 85.04 & 81.82 & 88.25 & 74.23 & 47.01 & 75.44 \\
Static-$\alpha$    & \textbf{79.55} & 60.64 & 89.33 & 80.36 & 64.93 & \textbf{89.07} & 86.03 & 85.86 & 71.13 & 47.01 & 75.39 \\
\midrule
\textbf{MG-Thinker} & 78.98 & \textbf{62.77} & \textbf{93.01} & \textbf{82.55} & 66.67 & 88.78 & \textbf{86.87} & 88.25 & 72.16 & \textbf{47.01} & \textbf{76.71} \\
\bottomrule
\end{tabular}
\end{adjustbox}
\end{table*}

\begin{table*}[t]
\centering
\caption{Performance on RefCOCO/+/g referring expression grounding. Best in \textbf{bold}, second-best \underline{underlined}.}
\label{tab:app-refcoco}
\begin{adjustbox}{max width=\textwidth}
\setlength{\tabcolsep}{4pt}
\begin{tabular}{l|ccc|ccc|cc|c}
\toprule
\multirow{2}{*}{\textbf{Models}} & \multicolumn{3}{c|}{\textbf{RefCOCO}} & \multicolumn{3}{c|}{\textbf{RefCOCO+}} & \multicolumn{2}{c|}{\textbf{RefCOCOg}} & \multirow{2}{*}{\textbf{AVG}} \\
\cmidrule{2-9}
& val & testA & testB & val & testA & testB & val & test & \\
\midrule
VisionLLM v2~\cite{wu2024visionllm} & 79.20 & 82.30 & 77.00 & 68.90 & 75.80 & 61.80 & 73.30 & 74.80 & 74.14 \\
Shikra & 87.00 & 90.60 & 80.20 & 81.60 & 87.40 & 72.10 & 82.30 & 82.20 & 82.97 \\
InternVL2-8B & 87.10 & 91.10 & 80.70 & 79.80 & 87.90 & 71.40 & 82.70 & 82.70 & 82.94 \\
GroundingGPT~\cite{li2024groundinggpt} & 88.02 & 91.55 & 82.47 & 81.61 & 87.18 & 73.18 & 81.67 & 81.99 & 83.57 \\
Griffon v2 & 89.60 & 91.80 & 86.50 & 81.90 & 85.50 & 76.20 & 85.00 & 86.00 & 85.30 \\
GroundingDINO-L~\cite{liu2024grounding} & 90.60 & 93.20 & 87.20 & 82.80 & 89.00 & 75.90 & 86.10 & 87.00 & 86.60 \\
Qwen2.5-VL-7B & 90.00 & 92.50 & 85.40 & 84.20 & 89.10 & 76.90 & 87.20 & 87.20 & 86.56 \\
Migician & 91.62 & 93.49 & 87.22 & \textbf{86.13} & \textbf{91.06} & 79.93 & 88.06 & 87.80 & 88.16 \\
UniVG-R1 & \textbf{91.64} & 93.11 & 87.16 & \underline{85.91} & \underline{90.53} & \textbf{80.04} & 88.67 & \textbf{88.56} & \textbf{88.20} \\
\textbf{MG-Thinker} & 91.28 & \textbf{93.55} & \textbf{87.48} & 85.79 & 90.50 & \underline{79.97} & \textbf{88.68} & \underline{88.22} & \underline{88.18} \\
\bottomrule
\end{tabular}%
\end{adjustbox}
\end{table*}
\begin{figure}
  \centering
  \includegraphics[width=0.85\linewidth]{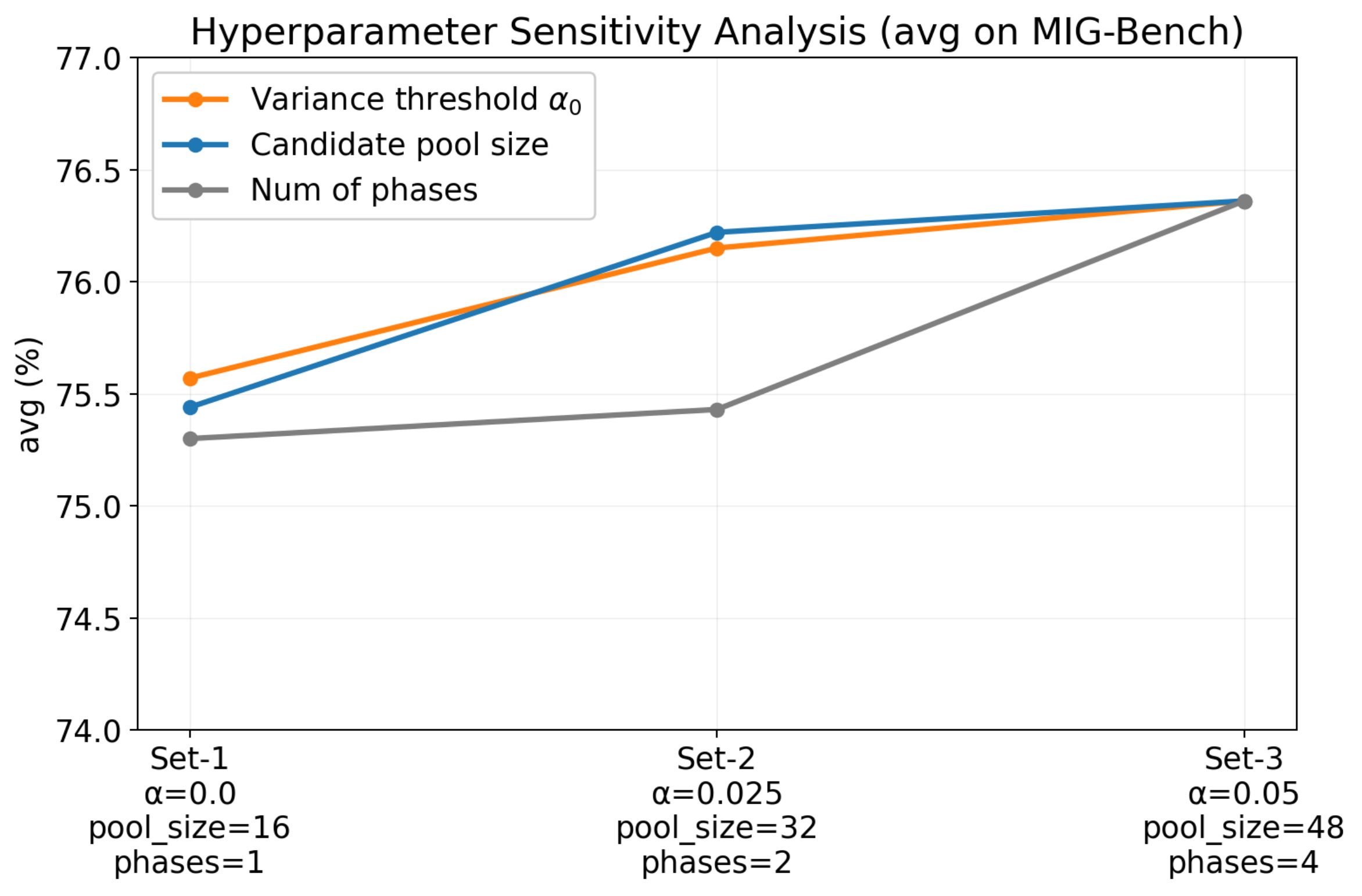}
  \caption{Continuous-variation sensitivity of BiA-DAPO on MIG-Bench: GIA threshold $\alpha_0$, candidate-pool size, and number of GIA phases.}
  \label{fig:app-sensitivity}
\end{figure}

\section{RL Post-Training Configuration and Hyperparameters}
\label{app:rl-config}
This section details the reinforcement post-training configuration (VeRL-based) and the BiA-DAPO-specific hyperparameters; basic settings (learning rate, hardware, global schedule) follow §4.1.
\begin{figure*}[t]
  \centering
  \includegraphics[width=1.0\linewidth]{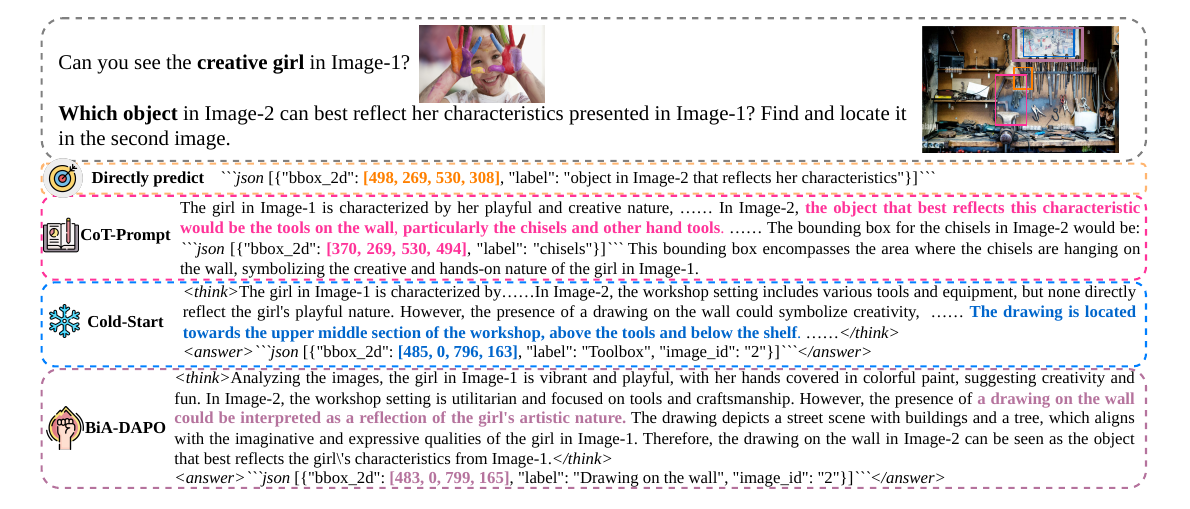}
  \caption{Comparison of non-reasoning and various reasoning paradigms on a representative cross-image reasoning grounding task: direct prediction, CoT-prompting, MG-Thinker cold-start, and MG-Thinker fully trained with BiA-DAPO post-training.}
  \label{fig:app-reasoning}
\end{figure*}
\begin{figure*}
  \centering
  \includegraphics[width=1.0\linewidth]{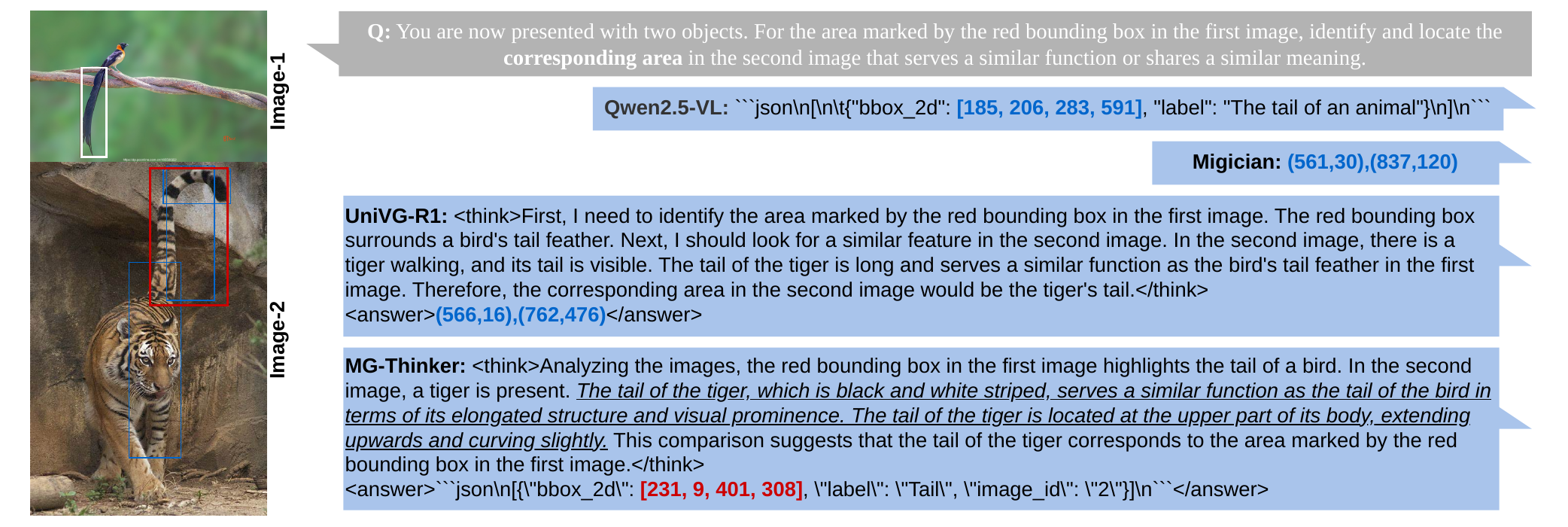}
  \caption{Visual analysis on MIG-Bench correspondence grounding tasks; ground-truth bounding-box annotations are shown in white to distinguish them from predictions.}
  \label{fig:app-case-correspond}
\end{figure*}
\noindent\textbf{VeRL-based RL configuration.} We implement RL post-training on top of the VeRL framework.\footnote{\url{https://github.com/verl-project/verl}} Tab.~\ref{tab:app-rl-settings} lists the RL-specific hyperparameters that most influence sampling behavior and stability. Inputs are loaded from RL-formatted parquet files keyed by \textit{prompt} and \textit{images}, with overlong-prompt filtering and strict truncation checks. We sample multiple responses per prompt during rollout and optimize the policy with the DAPO-style clipped surrogate (Appendix~\ref{app:dapo}) over a GRPO-style group-relative advantage estimator. KL regularization is applied as an explicit KL loss using a low-variance KL formulation (rather than a reward-shaping term).

\noindent\textbf{BiA-DAPO hyperparameters.} BiA-DAPO is instantiated by two complementary mechanisms, Group-Informativeness Assessment (GIA) and Cascaded-Reward Stratification (CRS), operating over a $3{\times}$ candidate pool. Three groups of hyperparameters are required: (i) a phase-decayed schedule with four phases for the GIA gate, (ii) the GIA threshold $\alpha_t$, and (iii) the candidate-pool size. The four-phase design originates from a one-epoch GRPO baseline scan: we collect the within-group rollout standard-deviation distribution and observe four well-separated regimes, zero-spread groups (sparsity-collapsed, no learning signal), low-, medium-, and high-spread groups, motivating a four-way partition aligned with these empirical boundaries. Following this categorization, we set the initial threshold to $\alpha_0{=}0.05$ and decay it by a factor of two at each subsequent phase, so that the gate progressively admits lower-spread groups as the policy stabilizes. For CRS, the candidate-pool size is fixed at $3{\times}$ the training batch (matching DAPO's generation batch) to ensure a consistent $\bar{r}_g$-ranking budget across update steps.

\section{Hyperparameter Sensitivity Analysis}
\label{app:sensitivity}
We further study BiA-DAPO's sensitivity on MIG-Bench by varying one hyperparameter at a time around the default configuration ($\alpha_0{=}0.05$, candidate pool $=48$, GIA phases $=4$). The continuous-variation curves are summarized in Fig.~\ref{fig:app-sensitivity}, while Tab.~\ref{tab:app-sensitivity-ablation} reports per-task accuracies at three representative single-axis corner cases. All averages in this section and in Fig.~\ref{fig:app-sensitivity} are computed under the Poll output format for every configuration, including the default, which therefore reads 76.36 here; Tab.~\ref{tab1} reports MG-Thinker's better-of-format result (76.71).

Specifically, varying the GIA threshold over $\alpha_0{\in}\{0.05,0.025,0.0\}$ yields average accuracies of $76.36/76.15\,(-0.21)/75.57\,(-0.79)$, indicating that a strictly positive threshold is beneficial and moderate variations are well tolerated; varying the candidate pool size over $\{48,32,16\}$ produces $76.36/76.22\,(-0.14)/75.44\,(-0.92)$, where an excessively small pool harms group-statistics stability while a medium pool stays close to the default; varying the number of GIA phases over $\{4,2,1\}$ gives $76.36/75.43\,(-0.93)/75.30\,(-1.06)$, showing that coarser schedules consistently underperform.

\noindent\textbf{Single-axis corner cases.} Tab.~\ref{tab:app-sensitivity-ablation} further examines three corner-case variants that each disable one BiA-DAPO design choice: \emph{Zero-Threshold} ($\alpha_0{=}0$, pool $=48$) opens the gate to all groups and is operationally equivalent to \emph{w/o.\ GIA}; \emph{Unexpanded Pool} ($1{\times}$ pool, i.e., pool $=16$) cancels the candidate-pool expansion so that group statistics are computed only over the training batch; \emph{Static-$\alpha$} ($\alpha{=}0.05$ frozen) freezes the threshold at its initial value without phase decay. All three consistently underperform the full default by $\sim$1 point on average and on most subtasks, confirming that each design choice contributes a non-redundant slice of the bi-axial mechanism. Overall, BiA-DAPO is robust under moderate variations but degrades under extreme settings, supporting the rationality of the chosen defaults.

\section{Analysis of Reasoning Paradigms (Case Study)}
\label{app:reasoning-cases}
Complementing the quantitative comparison in §4.2 \emph{Analysis of Reasoning Paradigms}, Fig.~\ref{fig:app-reasoning} provides a representative case on a cross-image correspondence task: identifying and localizing, in Image-2, the object that matches a target person in Image-1. Solving this task requires (1) \emph{anchoring} the target attributes in Image-1, (2) \emph{reasoning} over Image-2 candidates by attribute matching, and (3) \emph{grounding} the final decision via accurate localization.
\begin{figure*}
  \centering
  \includegraphics[width=1.0\linewidth]{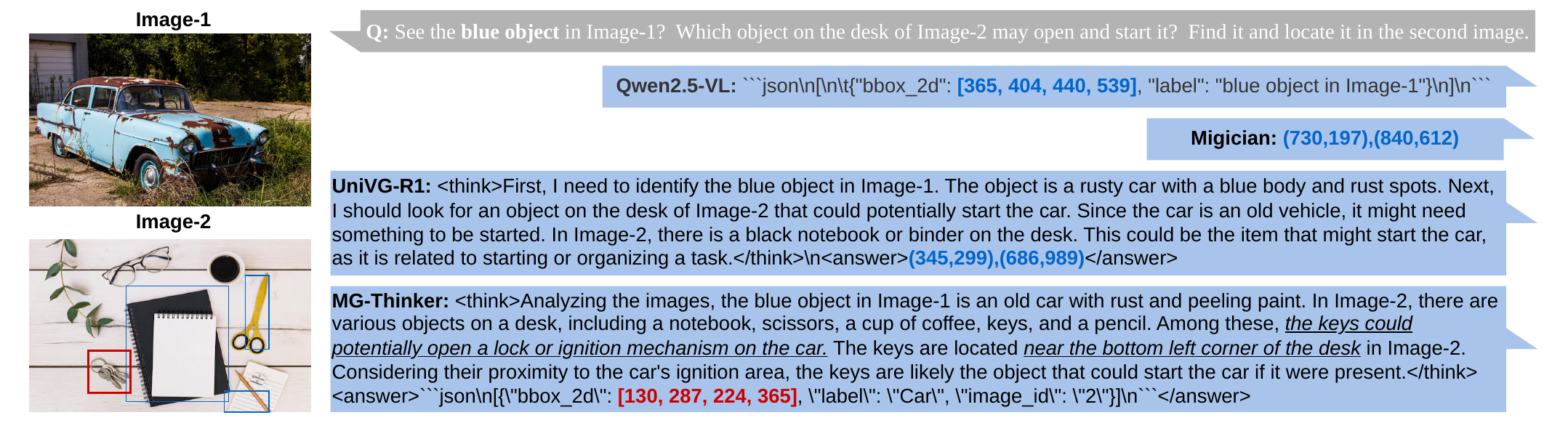}
  \caption{Visual analysis on MIG-Bench reasoning grounding tasks: comparison of MG-Thinker against representative multi-image baselines.}
  \label{fig:app-case-reason}
\end{figure*}
In the illustrated example, the target girl in Image-1 carries distinctive paint traces on her hands, face, and clothing, naturally suggesting drawing-related correspondences in Image-2. We compare four configurations:
\begin{itemize}
    \item \textbf{Qwen2.5-VL (direct prediction).} Without an explicit reasoning scaffold, the model fails to connect target attributes in Image-1 to candidates in Image-2, yielding incorrect localization.
    \item \textbf{Qwen2.5-VL\,+\,CoT prompting.} A generic CoT trigger induces a free-form rationale, but the trajectory may deviate from discriminative cues (\emph{reasoning drift}), still leading to erroneous localization.
    \item \textbf{MG-Thinker (cold-start).} The cold-start checkpoint can localize correctly but its rationale may be partially inconsistent with the visual evidence, indicating an underdeveloped reasoning--grounding alignment.
    \item \textbf{MG-Thinker (fully trained).} The BiA-DAPO post-trained model exhibits a systematic \emph{anchor--reason--ground} pattern: it first identifies target attributes in Image-1, then verifies matching cues in Image-2, and finally outputs accurate localization with a coherent explanation.
\end{itemize}

This case highlights distinct failure modes of non-reasoning inference and prompt-induced CoT, and demonstrates that our pipeline improves the \emph{consistency} between reasoning traces and grounding outputs, critical for reliable multi-image correspondence and localization.

\section{Performance on RefCOCO/+/g Benchmarks}
\label{app:refcoco}
We supplement the §4.2 \emph{General single-image grounding} discussion with the full RefCOCO/+/g comparison. Tab.~\ref{tab:app-refcoco} reports MG-Thinker against representative state-of-the-art single-image grounding methods, including VisionLLM v2~\cite{wu2024visionllm}, Shikra, InternVL2-8B, GroundingGPT~\cite{li2024groundinggpt}, Griffon v2, GroundingDINO-L~\cite{liu2024grounding}, Qwen2.5-VL-7B, Migician, and UniVG-R1. Despite being optimized for multi-image reasoning grounding, MG-Thinker remains strongly competitive on conventional single-image referring expression grounding, confirming that our post-training preserves general grounding capability rather than over-specializing to multi-image tasks.

\section{Qualitative Cases on MIG-Bench}
\label{app:qualitative}
Complementing the MIG-Bench quantitative results, we present qualitative case studies that contrast MG-Thinker with strong multi-image baselines, with an emphasis on dedicated multi-image grounding models. We analyze representative behaviors across semantic/logical reasoning, region selection under repeated distractors, and simultaneous multi-object tracking.

\noindent\textbf{Semantic and logical reasoning across image sequences.} Fig.~\ref{fig:app-case-reason} and Fig.~\ref{fig:app-case-correspond} examine cases requiring semantic logical reasoning and abstract cross-image association. Qwen2.5-VL under zero-shot prompting frequently fails to establish query-to-sequence correspondence, leading to inconsistent or arbitrary localization. Migician, while strong at direct end-to-end grounding, tends to underutilize semantic reasoning, resulting in rigid localization driven primarily by surface text cues. For reasoning-enabled baselines, reasoning can be a double-edged sword: UniVG-R1 may exhibit (i) plausible but incorrect reasoning that steers grounding toward wrong regions, or (ii) partially correct reasoning whose final localization is misaligned with the visual evidence, indicating a coupling gap between reasoning traces and spatial decisions. By contrast, MG-Thinker more reliably leverages semantic and logical constraints to regulate the reasoning process, and its reasoning signals translate into more precise and consistent localization.

\begin{figure*}
  \centering
  \includegraphics[width=1\linewidth]{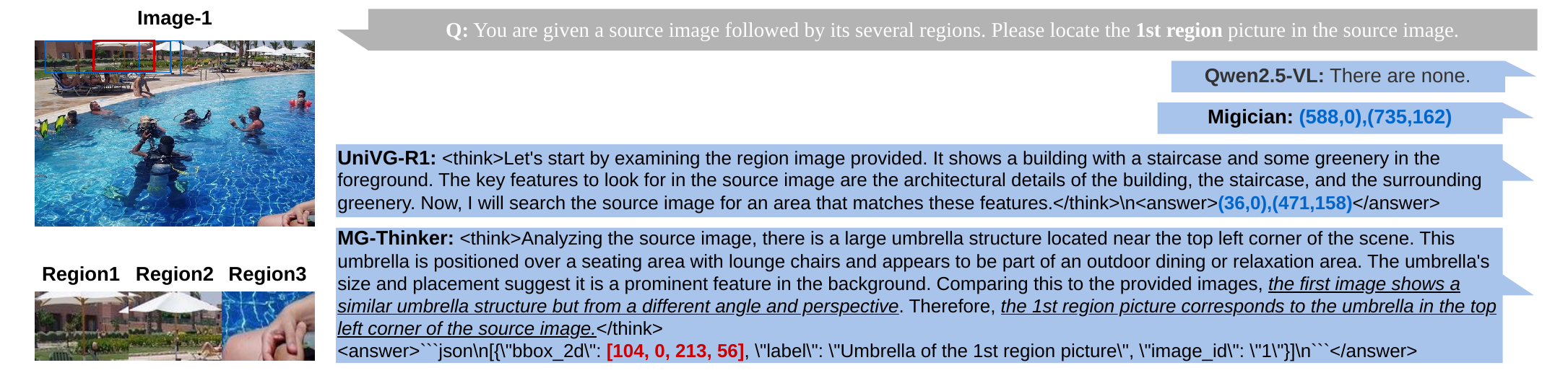}
  \caption{Visual analysis on MIG-Bench region grounding tasks under repeated distractors.}
  \label{fig:app-case-region}
\end{figure*}

\noindent\textbf{Multi-region selection under repeated distractors.} Fig.~\ref{fig:app-case-region} evaluates multi-image region grounding scenarios involving selection among multiple candidate regions and localization of inconspicuous, repeated small objects. MG-Thinker better suppresses interference from repeated instances by jointly considering positional relations and fine-grained visual attributes, yielding more stable region selection when distractors share similar appearance.

\begin{figure*}
  \centering
  \includegraphics[width=1\linewidth]{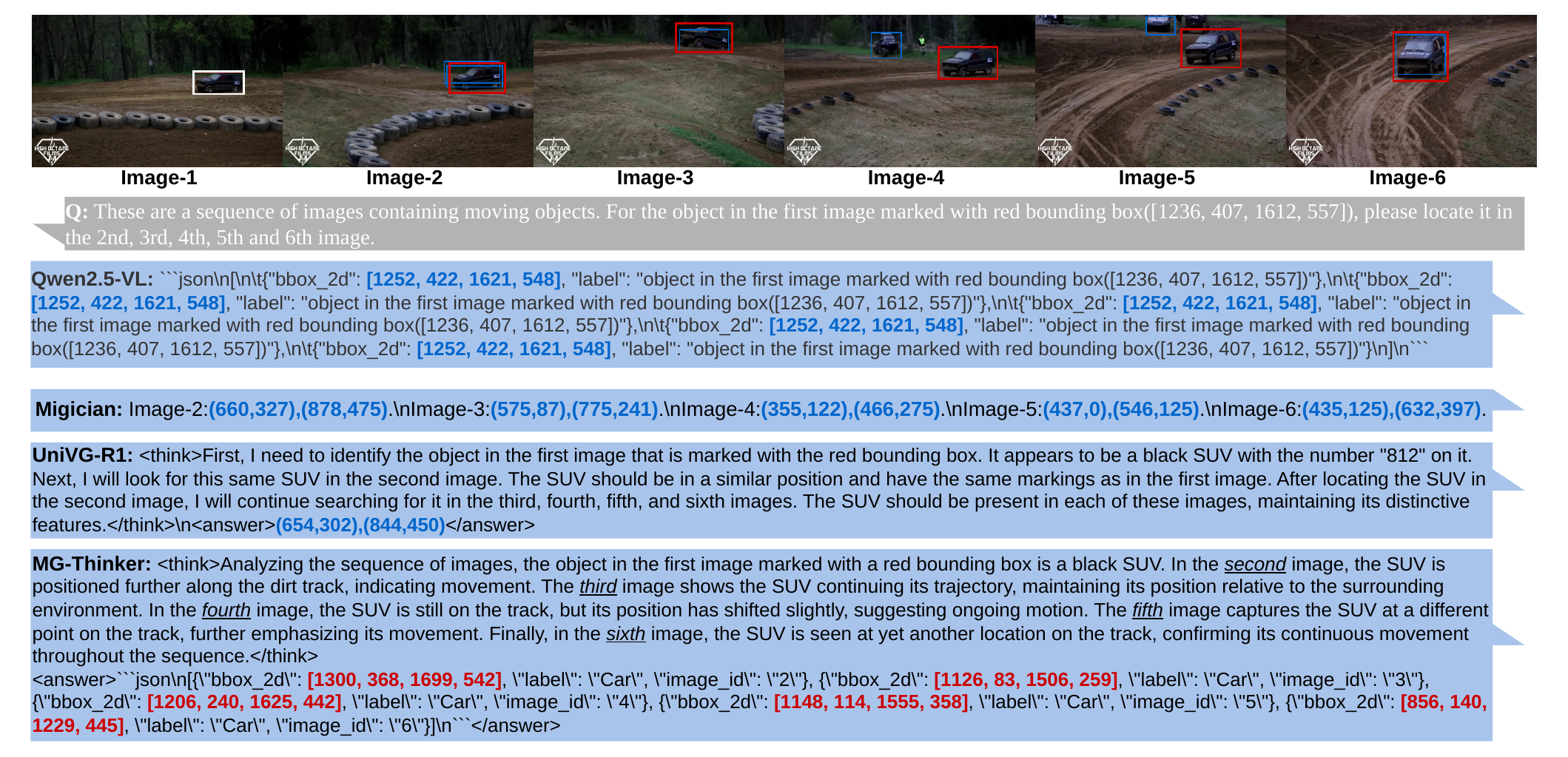}
  \caption{Visual analysis on MIG-Bench object-tracking grounding tasks; ground-truth annotations are shown in white. Numerical values follow each model's coordinate convention (relative or absolute).}
  \label{fig:app-case-track}
\end{figure*}

\noindent\textbf{Simultaneous multi-object tracking grounding.} Fig.~\ref{fig:app-case-track} highlights a challenging multi-object tracking setting where models must output multiple bounding boxes in a single inference. Qwen2.5-VL frequently produces chaotic, repetitive outputs; UniVG-R1 struggles to reconcile multi-object output with its reasoning-and-grounding procedure; and Migician's accuracy degrades when scaling to simultaneous multi-object localization. MG-Thinker remains more coherent in organizing outputs and maintaining alignment between predicted boxes and temporal/identity cues across frames, although the task remains challenging for all models.

\end{document}